\documentclass{ws-ijprai}

\usepackage{amssymb}
\usepackage{graphicx}
\usepackage{xcolor}

\usepackage{lineno}
\usepackage{subfigure}
\usepackage{amsmath}

\DeclareMathOperator*{\argminA}{arg\,min}

\begin{document}


%
\catchline{}{}{}{}{}
%


\title{A Search for the Underlying Equation Governing Similar Systems}

\author{Changwei Loh}

\address{Siemens Ltd., China Corporate Technology (Suzhou), \,\\
18 Jinfang Road, Dongfang Innovation Park, Suzhou, Jiangsu, China\,\\
\email{changwei.loh@siemens.com}
}

\author{Daniel Schneegass}

\address{Siemens Ltd., China Corporate Technology (Beijing),\,\\
17 Wangjing Zhonghuan Nanlu, Beijing, China\,\\
\email{daniel.schneegass@siemens.com}
}

\author{Pengwei Tian}

\address{Siemens Ltd., China Corporate Technology (Beijing),\,\\
17 Wangjing Zhonghuan Nanlu, Beijing, China\,\\
\email{pengwei.tian@siemens.com}
}

\maketitle


\begin{abstract}
We show a data-driven approach to discover the underlying structural form of the mathematical equation governing the dynamics of multiple but similar systems induced by the same mechanisms. This approach hinges on theories that we lay out involving arguments based on the nature of physical systems. In the same vein, we also introduce a metric to search for the best candidate equation using the datasets generated from the systems. This approach involves symbolic regression by means of genetic programming and regressions to compute the strength of the interplay between the extrinsic parameters in a candidate equation. We relate these extrinsic parameters to the hidden properties of the data-generating systems. The behavior of a new similar system can be predicted easily by utilizing the discovered structural form of the general equation. As illustrations, we apply the approach to identify candidate structural forms of the underlying equation governing two cases: the changes in a sensor measurement of degrading engines; and the search for the governing equation of systems with known variations of an intrinsic parameter.
\end{abstract}


\section{Introduction}
\label{intro}

Scientists and engineeers have always been keen in seeking valuable insights about a system and hidden patterns from its empirical data. The culmination of such laborious work is when the governing equation is found, thus discovering the essence of the system and thereby advances related technologies. The availability of big data in recent years offers opportunities of adopting multiple approaches to such discoveries.

The long known symbolic regression (SR) via genetic programming (GP)~\cite{Koza} (SR) presents one such approach. GP is a biologically inspired machine learning method which can be used to find a mathematical equation that best describes the observed data from a system. By concatenating primitive building blocks, SR constructs possible equations spanned by these building blocks. It probabilistically reconstitutes and cross-combines previous equations to form new ones in the hope of finding a better solution to describe the data. The dilemma of finding a model -- in the form of an equation -- that explains the data is that there are at least two criteria to be optimized, and that the best model is supposed to satisfy these two criteria: (1) fits the data well, and (2) compact (in the spirit of parsimonisity). Data fitting is a common theme in data analysis with various fitting methods to choose from. However, the parsimonisity (Occam's Razor) criterion -- in the sense that a "simple" model is favored rather than a more "complicated" model -- has had different interpretations~\cite{Schmidt81,10.1007/978-3-642-27549-4_34,inbook,4632147}. What do we mean by being simple? Do we regard a model having multiple terms or multiple variables to be complicated? If so, how simple can a model be and yet still satisfies criteria (1)? Or, should parsimonisity be interpreted differently? It is a challenge to identify the best model based on multiple criteria. In particular, satisfying both Criteria (1) and (2) implies that there could be more than one model compatible with the data since it is generally impossible to simultaneously optimize on both criteria. The minimum description length (MDL) principle~\cite{MDL,Grnwald2004ATI} also attempts to find a solution that satisfies both Criteria (1) and (2). In MDL, each model is viewed as a code (for example, in units of bits), and the code length needed to describe the model measures its complexity, in line with Occam's Razor. Of course, the model with the smallest code length as selected by MDL also needs to adequately describes the data apart from being simple. The MDL criterion offers a solution to the bias-variance dilemma~\cite{BV}, i.e. finding the balance between the complexity of a model and its goodness of fit to the data. However, the definition of a code remains elusive.

In this paper, we attempt to study this problem for situations where there are an abundance of data from multiple but similar systems experiencing the same dynamics. In such situations, we show that there is a general way to extract the underlying mathematical equation, in particular its structural form governing the dynamics by leveraging the collective information from these systems. In doing so, we establish a different meaning to a compact model (parsimonisity) on the basis of the number of hidden system properties that may be taking part in the dynamics. For model comparisons, we introduce a metric -- taking into account both Criteria (1) and (2) -- using arguments pertaining to the nature of physical systems. Learning about a system through the knowledge gained from other similar systems is an idea that has been successfully used in transfer learning~\cite{Weiss2016}, where a neural network model is first trained on data samples generated from other systems, and the model is then refined with data generated from the systems of interest to an analyst. However, transfer learning differs from our work in two major aspects: (1) similar systems as considered in our work are those that experience the same physical dynamics or operating conditions; and (2) we construct an interpretable model in the form of an equation w.r.t the data-generating systems instead of a black-box model.

The paper is structured as follows: In Section~\ref{symbolic}, we digress from our main work to give a brief overview of SR via GP; in Section~\ref{theory}, we present the theories of our work; in Section~\ref{approach}, we propose an approach based on the theory to search for the underlying equations governing the dynamics of multiple systems; and finally, in Section~\ref{application}, we give two illustrative applications of our work.

\section{Overview: Symbolic Regression via Genetic Programming}
\label{symbolic}

\begin{figure}
  \centering
    \includegraphics[width=0.6\textwidth]{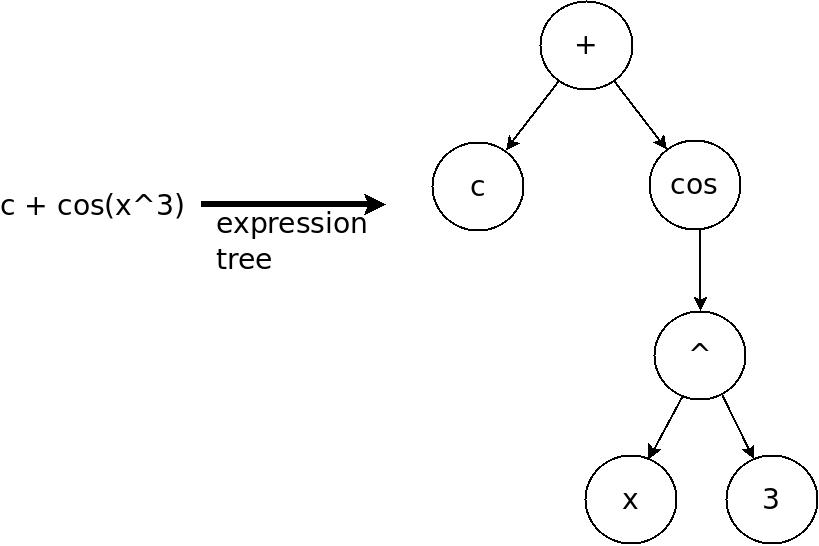}
  \caption{An expression tree for $c + \cos(x^3)$.}
  \label{example}
\end{figure}

\begin{figure}
  \centering
    \includegraphics[width=0.5\textwidth]{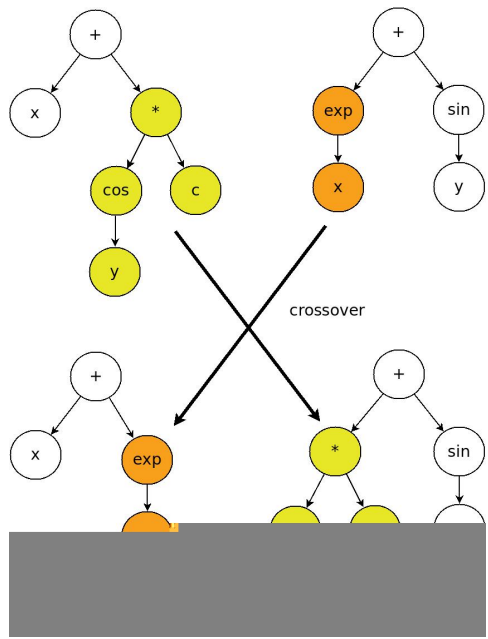}
  \caption{Crossover process in genetic programming. }
  \label{crossover}
\end{figure}

\begin{figure}
  \centering
    \includegraphics[width=0.6\textwidth]{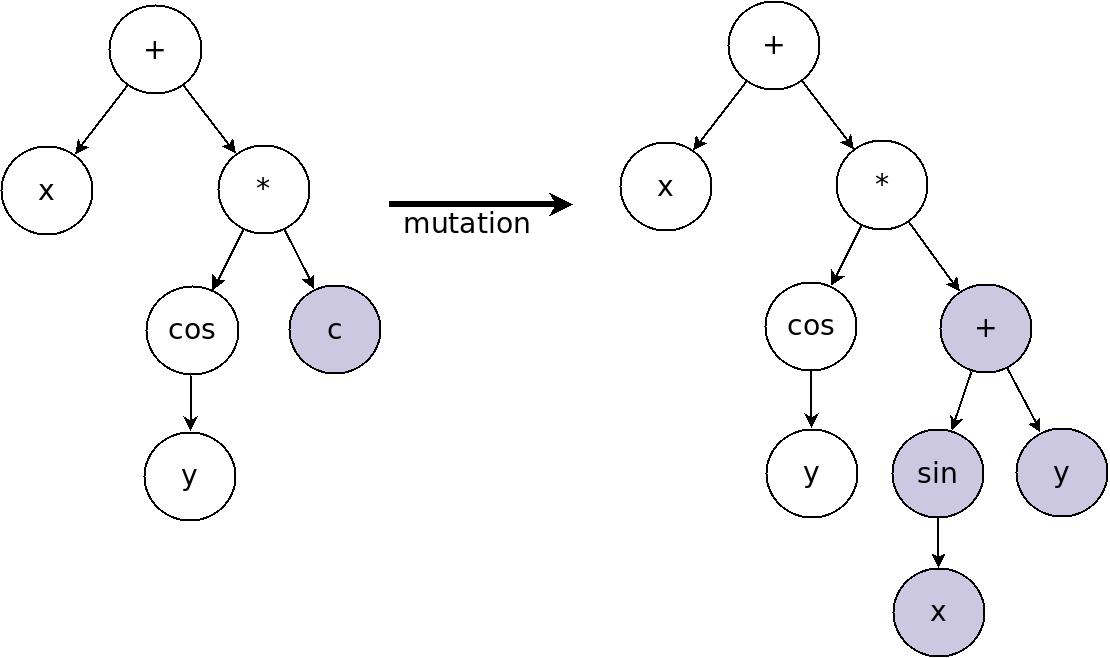}
  \caption{Mutation process in genetic programming. }
  \label{mutation}
\end{figure}

An equation can be represented as an expression tree (for example, see Figure~\ref{example}) comprising of a concatenation of nodes. Each node in the tree constitutes a primitive building block of the expression: an algebraic operator (e.g., $+$, $-$, $\times$, $\div$), an analytical operator (e.g.,  $\exp(.), \log(.)$ and $\cos(.)$), a constant or a variable~\cite{Koza, Schmidt81, CHEN20181973}.
In genetic programming, there are additional genetic operations, for example crossovers and mutations that can randomly act on subtrees to form new equations from old ones, thus is said to have evolved an equation from the old generation to a new equation. In a crossover, two subtrees from different trees can be swapped to give two new trees. For example, in Figure~\ref{crossover}, the equations $x + c \cos y$ and $e^x + \sin y$ are crossovered to give two new equations, i.e. $x + e^x$ and $c \cos y + \sin y$. In a mutation, a subtree can be altered, hence evolving into a new tree. For example, in Figure~\ref{mutation}, the equation $x + c\cos y$ is mutated to $x + (\sin x + y)\cos y$.

The aim of symbolic regression is to find a mathematical model -- in the form of an equation -- to a data-generating problem. It explores the function space and generates candidate equations based on a set of pre-determined primitive building blocks. By evolving equations using genetic programming, equations from the new generation is expected to be better in explaining the data compared to equations in the old generation. The measure of comparison is known as a fitness metric which usually involves computing the mean square loss between the data and the model predictions. The best candidate equation(s) -- in terms of the fitness value -- from a generation is usually retained to form part of the next generation, while other equations are mutated and/or crossovered to form new ones. 



\section{Theory}
\label{theory}

We consider situations where multiple datasets are available. Ideally, we would like to have an ensemble of independent but identical systems, each generating a set of data. The average fitness is then used to compare candidate equations when searching for the one that best explains the dynamics of the systems. At the very least, we hope that we could repeat an experiment on the same system (without destruction) multiple times to obtain several datasets. However, in practice especially in non-laboratory settings, such ensembles can hardly be found. Instead, there could be multiple but similar systems, each experiencing the same dynamics. In this Section, we present a way to combine these multiple but similar systems, so that that their collective information can be used to find the underlying equation common to them all. In doing so, we show that the parsimonisity of the model can be computed through an entropy metric, which relates to the number of independent hidden properties of the systems that could have taken part in the dynamics.

The understanding that the collected data are not purely mathematical construct, but are measurements on physical entities gives rise to two conditions when building a mathematical relationship: (1) the numerical values as computed from both sides of the equation should clearly be equal; and, (2) the dimension (physical units) of both sides of the equation should be consistent. This second condition is usually of less importance during data analyses but will play an important role here. Dimension consistencies has been used in~\cite{ecology} to some extent. Here, we explore the need for dimension consistency in detail.

We set the stage with a hypothetical system instrumented with some measurement devices to enable data collection. We wish to find a mathematical relationship between the variables $\{x, y\}$ related to the data collected, for instance temperature and density. We assume $x \in \mathbb{R}$ and $y \in \mathbb{R}$ as the independent and dependent variables respectively. The arguments herein can easily be generalized to multiple dependent variables. 

Suppose an equation contains analytical operators, for instance $\mathbb{\sin(.)}$, $\cos(.)$ and $\log(.)$, then in order for the dimension of both sides of any equation to be consistent, each analytical operator $O$ can only exist as a product of an unknown dimensionful coefficient $\gamma$ and the operator $O$ itself, i.e. $\gamma O$.  Moreover, since the physical units of the variables are generally different, the equation that relates $y$ and $x$ should admit a composite of the form: $\alpha (x + x_0) \equiv \alpha x + \beta$ instead of a "bare" $x$; where $\alpha$ and $\beta$ are dimensionful unknowns. This same form applies even if the variables appear as operands to an analytical operator.  The constant $x_0$ amounts to a shift in the coordinate system which has no impact on the structural form of an equation, and in practice, can be interpreted as accommodating any non-zero offsets/calibrations of the measurements in $x$.

Note that, we regard an exponentiation $x^n$ as an operator, hence its proper form should be $\gamma (\alpha x + \beta)^n$. The reason can be understood from the Taylor expansion of an analytical operation $O(x) = a_0 + a_1 x + a_2 x^2 + ... + a_n x^n + ...$; when transforming the operator $O(x)$ to $\gamma O(\alpha x + \beta)$, the Taylor expansion will be of the form $ a_0 \gamma + a_1 \gamma (\alpha x + \beta) + ... + a_n \gamma (\alpha x + \beta)^n + ...$

For the purpose of this work, we will refer $\alpha$, $\beta$ and $\gamma$ as external parameters. As an illustration, we show in Equation~\ref{invalid}, an invalid hypothesis due to the dimensions:
\begin{equation}
\label{invalid}
y = x^n + e^{x}\cos x.
\end{equation}
For dimension consistency, we transform the hypothesis to a proper form as shown in \ref{valid}, where both sides of the equation are dimensionless:
\begin{equation}
\label{valid}
my + y_0 = \gamma_0(\alpha_0 x + \beta_0)^n + \gamma_1 e^{(\alpha_1 x + \beta_1)} \gamma_2 \cos(\alpha_2 x + \beta_2),
\end{equation}
where $m$, $y_0$, $\alpha_i$, $\beta_i$ and $\gamma_i$ are some unknown external parameters.

By rearranging and redefining the external parameters, we have the simplified equation:
\begin{equation}
y = \gamma_0(\alpha_0 x + \beta_0)^n + \gamma_1 e^{\alpha_1 x}\cos(\alpha_2 x + \beta_2) + y_0.
\end{equation}
%

With the external parameters stated explicitly, an equation that relates $y$ as a function of $x$ can be written as
\begin{equation} 
y = f(x ; y_0, \alpha_0,\alpha_1...,\alpha_p,\beta_0,\beta_1,...,\beta_q,\gamma_0,\gamma_1,...,\gamma_r).
\end{equation}
The dynamics of similar systems induced by the same mechanisms, actions or operating conditions are reflected in the function $f$. The extrinsic parameters are merely combinations of the hidden properties (intrinsic parameters) of the systems that took part in the dynamics as exhibited by the structural form of $f$. In other words, the values of the external parameters can be different for each system thus describing uniquely the different magnitudes of the intrinsic parameters, but the structural form of the function $f$ is a global feature of the set of similar systems. In this regard, we interpret the systems as different curves lying on the same manifold embedded in a higher dimensional space -- from the perspective of the dynamics -- parameterized by the extrinsic parameters and the independent variable $x$. We refer readers elsewhere for further readings on the notion of data manifolds~\cite{Lei2018GeometricUO,Lin2015,inproceedings}. The following examples concretely illustrate the notion of similar systems:

\begin{itemize}
\item in studying velocities of pendulums, swinging pendulums with varying lengths of the string attached to different massive bobs are considered similar systems; a swinging pendulum with a spring attached to a bob is not similar because Hooke's law~\cite{Mwanje_1980} comes into play due to the elastic spring.

\item in studying property prices, houses in various regions with a sea view are considered similar systems; houses in a region with a river view is not similar.
\end{itemize}

Denote $m$ as a candidate model that attempts to explain some given datasets, we propose that the best model $m^*$ should be the model that satisfies two Conditions: (1) fits the data well, and (2) invokes minimal number of independent system properties to explain the data. With respect to Condition (2), this means we advocate that the number of independent hidden properties is less than the number of extrinsic parameters $\Theta$. This implies that (1) some of the parameters $t$ could be numerical constants with respect to the given datasets, and/or (2) $\exists$ $\theta_k \in \{\theta_1,...,\theta_{\Theta}\}$ $|$ $ \theta_k = g(\theta_1,...,\theta_{k-1},\theta_{k+1},...,\theta_{\Theta})$ for some function $g$. Let $L$ be a metric involving entropies for continuous variables defined as follows:
\begin{equation}
L = KL(S,m) + h(\theta_1,\theta_2,...,\theta_{\Theta}),
\end{equation}
where $KL(S,m)$ is the Kullback-Leibler divergence between the systems (the empirical data distributions) and the candidate model which measures how well the model approximates the data; and $h(\theta_1,\theta_2,...,\theta_{\Theta})$ is the joint entropy of the distributions of all $\Theta$ extrinsic parameters in the model which measures the strength of the relationships between the parameters; 
then, the best model $m^*$ is the one such that:
\begin{equation}
\label{mstar}
m^* = \argminA_{m} L.
\end{equation}
The first and second term in $L$ embodies Condition (1) and (2) respectively.
The joint entropy can be further expressed as a summation of conditional entropies:
\begin{equation}
h(\theta_1,\theta_2,...,\theta_{\Theta}) = \sum_{k=1}^{\Theta} h(\theta_k | \theta_{k-1},...,\theta_1).
\end{equation}
If a parameter $\theta$ is a constant over the datasets (disregarding noise in the data), the conditional entropy $h(\theta|.) = 0$ implying that only "true" extrinsic parameters play a role in the joint entropy. A constant extrinsic parameter also indicates that it is at most, dependent on some non-varying hidden property of the systems currently being considered, which means it effectively -- from the perspective of these systems -- does not participate in the dynamics. Moreover, suppose some $\theta_j$ is some function of another parameter $\theta_i$ in the model, then $h(\theta_j|\theta_i) = 0$. Hence, $\theta_j$ also drops out from the joint entropy. 

Thus, the joint entropy between the extrinsic parameters indirectly measures the number of varying independent hidden properties of the systems that generate the datasets. Consequently, we view the parsimony measure or complexity of the model to be measurable using the joint entropy. As a consequence, a model can be complex (with many terms, or an expression tree that can be deep with many nodes), and yet is still a simple model. Note, although polynomial models could have low Kullback-Leibler divergence, the relationships between the extrinsic parameters in a polynomial may be weak, i.e. high joint entropy.

Let $D$ be the number of similar systems generating in total $D$ number of datasets, where the $j^\mathrm{th}$ dataset contains $d_j$ number of data points, then finding $m^*$ through the minimization of the $L$ metric -- in terms of the empirical data -- is equivalent to minimizing the following alternative form of $L$:
\begin{equation}
\label{simplemstar}
\begin{aligned}
L &= \sum_{j=1}^{D}\sum_{i=1}^{d_j} \frac{1}{Dd_j} (y_{ij}^{data} - y^{pred}_{ij})^2 + \sum_{k=1}^{\Theta}\sum_{j=1}^{D} \frac{1}{D}  (\theta_{jk}^{data} - \theta_{jk}^{pred} )^2  \\
 &\equiv L_1 + L_2,
\end{aligned}
\end{equation}
where $\theta_i^{pred} = \theta_i^{pred}(\theta_1,\theta_2,...,\theta_{i-1})$, i.e. the predictions on $\theta_i$ based on information from $\theta_1,...,\theta_{i-1}$. The KL-divergence in $L_1$ and conditional entropies in $L_2$ are monotonically related to the first and second mean square loss terms~\cite{Goodfellow-et-al-2016, FRENAY20131} in Equation~\ref{simplemstar} respectively. Note that, $\theta_1^{pred}$ is a single value equal to the mean of the $\theta_1$ data points since there are no other information left. Moreover, when there is only one dataset, i.e. $D=1$, Equation~\ref{simplemstar} reduces to
\begin{equation}
L =  \sum_{i=1}^{d_1} \frac{1}{d_1} (y_{i}^{data} - y^{pred}_{i})^2, 
\end{equation}
since all extrinsic parameters in a candidate equation can only be treated as constants from the perspective of a single dataset. On the other hand, suppose all $D$ datasets were generated from one single system, i.e. constant intrinsic parameters, then the $L$ metric can also be viewed as the average $L$ of that system.

In computing these entropies, the candidate equation should preferably be in its simplest form, i.e. the parameters have been combined and simplified to an irreducible form before the computation. Otherwise, the computed entropy would be larger than expected. As an example, suppose a system is characterized by a constant $c$, but the model has $\theta_1$ and $\theta_2$, with $c = \theta_1 + \theta_2$, then the following is a contradiction:
\begin{equation}
h(\theta_1,\theta_2) = h(\theta_1 | \theta_2) + h(\theta_2) = h(\theta_2) > 0,
\end{equation}
but the system has zero entropy since $h(c) = 0$ by virtue of it being a constant.

By combining the notion of dimension consistency and Equation~\ref{simplemstar}, one is now equipped for finding the common underlying equation that best describes the dynamics acting upon the multiple systems. Although this Section is grounded in arguments based on the nature of physical systems, it is worth mentioning that, the transformations based on dimension consistencies can also be used with data from purely mathematical construct. For the $L$ metric, since its second term $L_2$ is based entirely on physical arguments (Condition (2)), it is less useful for pure mathematical data.

\section{Approach}
\label{approach}

In this Section, we propose an approach in line with the theories as laid out in Section~\ref{theory}.
We again denote $D$ as the number of given datasets. Our approach starts by utilizing SR with GP to find candidate equations. For each candidate equation generated using GP, we transform the candidate using the following set of transformations to insert the necessary extrinsic parameters:
\begin{equation}
\begin{aligned}
\textrm{independent variable, } x &\rightarrow \alpha x + \beta \\ 
\textrm{analytical operator, } O &\rightarrow \gamma O \\ 
\textrm{candidate equation} &\rightarrow \textrm{candidate equation } + y_0 \\
\end{aligned}
\end{equation}
Sometimes, GP will propose a candidate containing numerical constants. In this case, we replace the numerical constants with additional unknown extrinsic parameters. In total, there will be $T$ number of extrinsic parameters associated with each dataset for a given candidate. 
Incidentally, the transformation procedure has an added benefit for any generic SR with GP search:
one candidate equation from GP is mapped to many candidates after the transformation, hence speeding up GP searches by reducing the search space.

With all the extrinsic parameters inserted, in principle we should proceed with a simultaneous fit to all the datasets to obtain the smallest $L$ metric value associated with the candidate, thereby obtaining the best fitted values for each extrinsic parameters, i.e. $D\Theta$ parameters. The expensive computation is reflected in the interplay between all $D\Theta$ parameters in the second term ($L_2$) of $L$. Here, we propose an alternative assuming the impact is minimal. In the alternative method: (1) we first use GP search with the fitness metric defined by $L_1$; (2) then we proceed with computing $L_2$ for some candidates with low $L_1$; (3) then finally we compute $L = L_1 + L_2$ and regard the best candidate as the one with the lowest $L$.

In detail, the first step in the alternative method is a follows:
we do a curve fitting process on a dataset with the transformed candidate to find the minimum value to the first term ($L_1$) in $L$, thereby indirectly obtaining a vector containing $T$ values, where each value is asscociated with one of the parameters. This means we are finding the best $T$ values with the transformed candidate that approximates that dataset. We repeat the curve fitting process for each of the $D$ datasets. 
At the conclusion of the curve fitting processes, we should have $D$ vectors of numerical values, with each vector having $T$ elements. Any curve fitting methods can be used. Here, we use the Levenberg-Marquardt algorithm~\cite{toush} and a particle swarm optimization (PSO)~\cite{MARINI2015153} as our curve fitting methods. The vector of $T$ values are picked from whichever method that gives the lower $L_1$ value. The Levenberg-Marquardt algorithm is a common method used for finding solutions to non-linear least square problems, utilizing gradient descent and Gauss-Newton methods in the process. In PSO, the solution to the curve fit is structured as an optimization problem. This machine learning algorithm mimics a group of social birds, each with some velocity to visit different points in the mathematical search space and searches for the minima (solution to the problem) through their intra-group communications of the visited points and their associated value of $L_1$; where the birds tend to move towards the lower $L_1$ as found by any one bird at each iteration.  After performing all $D$ curve fits, we compute the mean of $L_1$ and use this as the fitness value associated with the candidate equation for GP to search for the next generation of new candidates. We retain the best old candidate as part of the new generation for the GP search. In addition, this candidate and its corresponding $DT$ parameters are recorded for the next step.

After several generations of GP search, we should have a list of candidate records. The following should be performed for each candidate separately:
with the $D$ vectors for a candidate, we proceed with evaluating the second term ($L_2$) in $L$. We first divide the vector samples into a training set and a test set. We use a regression-type machine learning (ML) algorithm like Decision Tree Regressor~\cite{doi:10.1002/widm.8} to train the ML model on the training set consisting of some fitted values of each parameter $t_i$ given the fitted values of the other parameters $\theta_1,...,\theta_{i-1}$. Specifically, the regressor has $i-1$ inputs and one output, i.e. $\theta_i$. Once trained, the samples in the test set is used to compute the loss function of the ML model, i.e. the mean square error loss $MSE$: 
\begin{equation}
MSE = \frac{1}{N_{\textrm{test set}}}\sum (\theta^{data}_i - \theta^{pred}_i(\theta^{data}_1,...,\theta^{data}_{i-1}))^2,
\end{equation}
where the summation is over the number of samples in the test set. We repeat the training of Decision Tree Regressors for $\Theta-1$ times corresponding to predicting the fitted values for $\theta_2,\theta_3,..,\theta_{\Theta}$. For $\theta_1$, we can simply calculate $\theta_1^{pred}$ as the mean of the fitted values of $\theta_1$. The summations of the $MSE$ is $L_2$. At the conclusion of this step, we sum $L_1$ and $L_2$ to obtain $L$ of the candidate. The best equation is the one having the lowest $L$. Due to the nature of this alternative method, the best equation usually has a low $L_1$, which in any case, should be expected for any candidate that fits the data.

When doing curve fittings and regression model trainings, it is common to perform normalizations to speed up the computations. Note that the normalization constant of each dataset is most likely not the same. Therefore, the results should be unnormalized before calculating the $L$ metric. This would then be consistent with the notion that the datasets are generated from multiple but similar systems that lie on the same manifold.

In the next Section, we apply the approach as presented in this Section to find the underlying equations for two illustrative applications.

\section{Illustrative Applications}
\label{application}

\subsection{Case: sensor measurements of turbofan engines}
\label{nasa}

The dataset used in this case is publicly available at~\cite{NASA}  which contains simulation of the degree of degradation of turbofan engines per cycle $c$. Various sensor channels were recorded for the simulation. Here we used one of the sensor channel measurements denoted by $S$ as an illustration.

\begin{figure}
  \centering
  \subfigure[] {
    \includegraphics[width=1.\textwidth]{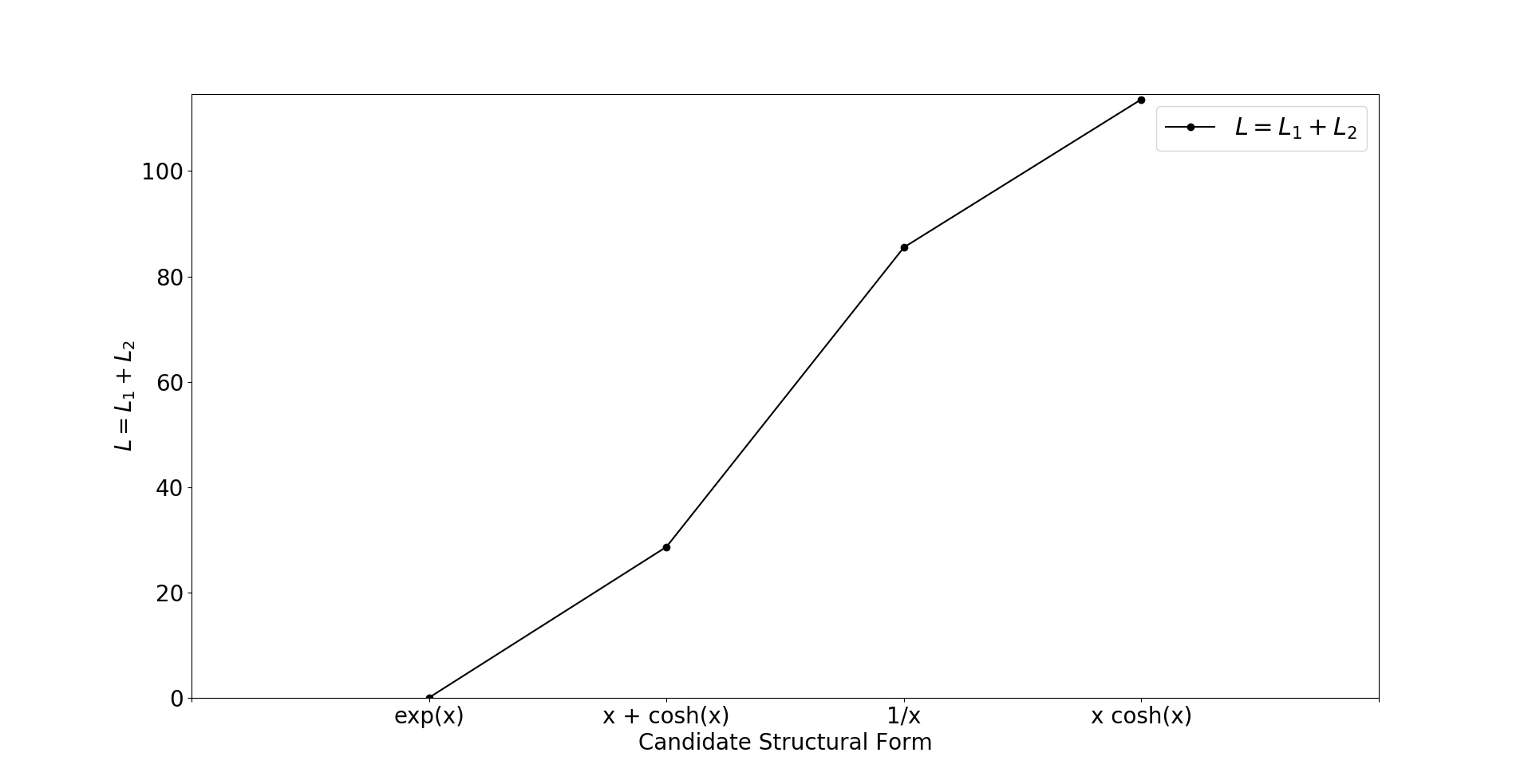}
    \label{fig_1t}
  }
  \subfigure[] {
    \includegraphics[width=1.\textwidth]{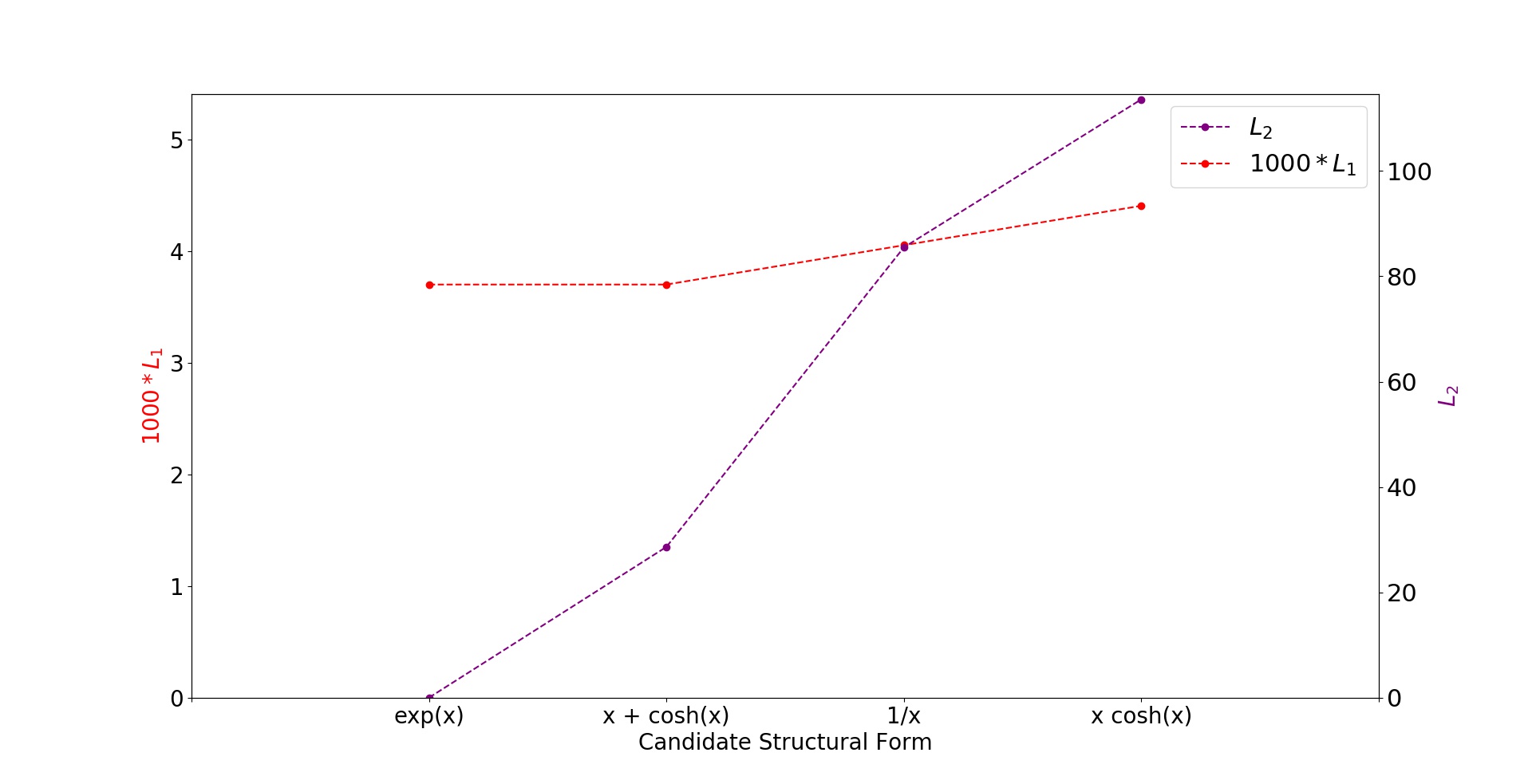}
    \label{fig_2t}
  }
  \caption{Structural form candidates (before transformation) with low values for the $L$ metric (see Figure (a)). $L_1$ and $L_2$ (see Figure (b)) are the values for the first and second terms in Equation~\ref{simplemstar}.}
  \label{turbofan_metric}
\end{figure}

\begin{figure}
  \centering
  \subfigure[] {
    \includegraphics[width=0.47\textwidth]{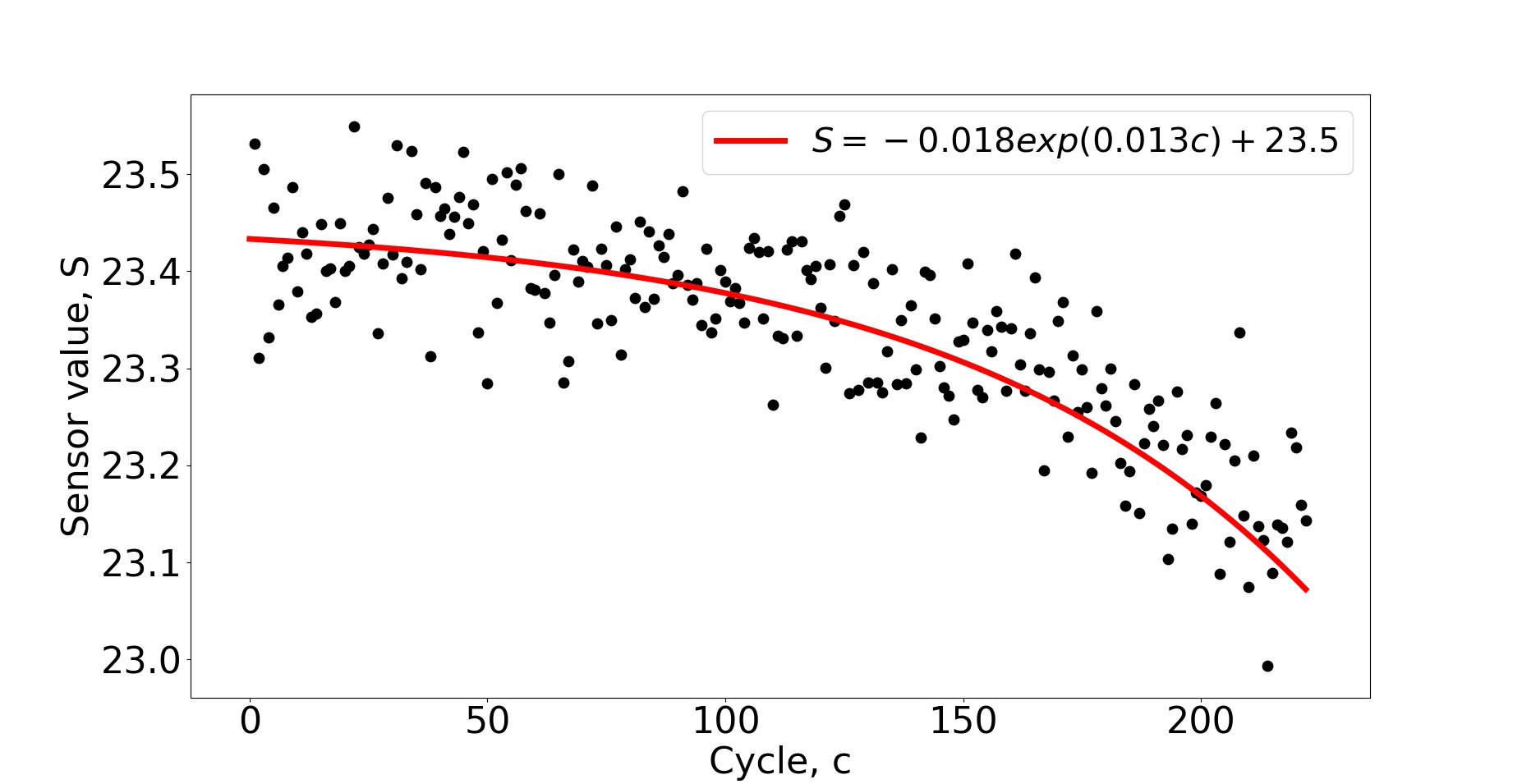}
    \label{fig_1}
  }
  \subfigure[] {
    \includegraphics[width=0.47\textwidth]{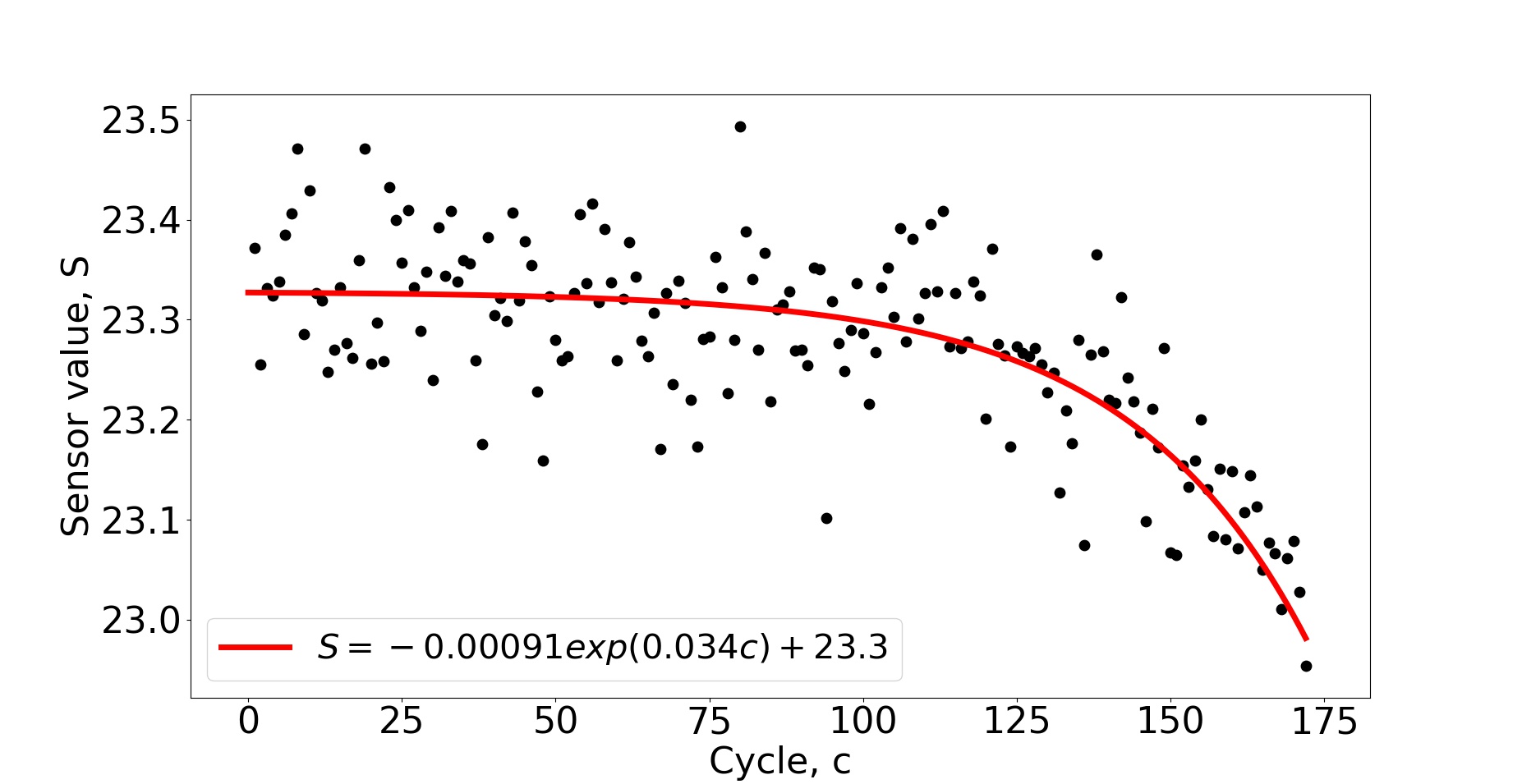}
    \label{fig_2}
  }
  \subfigure[] {
    \includegraphics[width=0.47\textwidth]{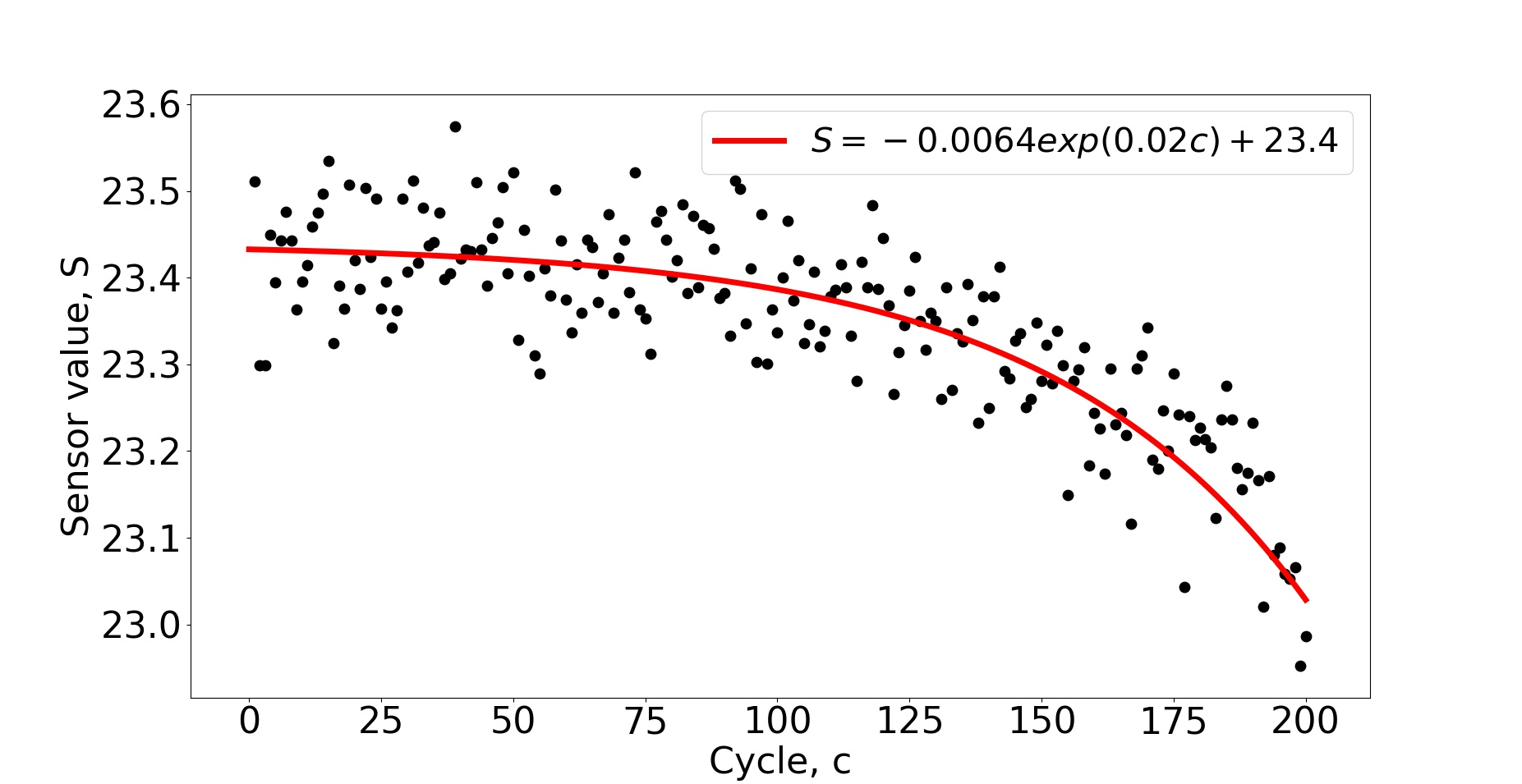}
    \label{fig_3}
  }
  \subfigure[] {
    \includegraphics[width=0.47\textwidth]{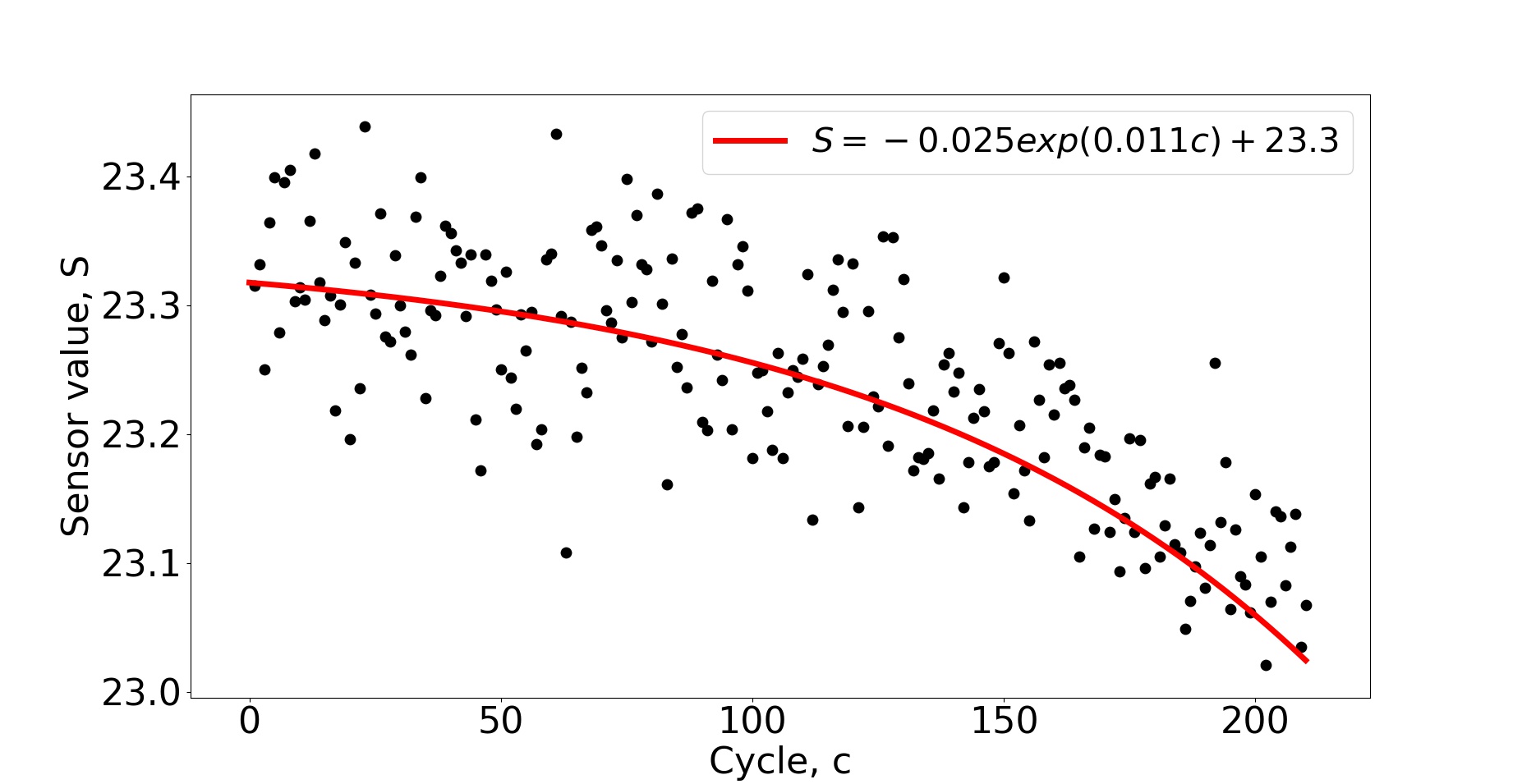}
    \label{fig_4}
  }
  \caption{Data points of four example turbofan engines and their fitted curves using the found structural form of the general equation. We report the value for $\gamma_0 e^{\beta_0}$ instead of the individual values for $\gamma_0$ and $\beta_0$. }
  \label{engines}
\end{figure}

\begin{figure}
  \centering
  \subfigure[$\frac{\gamma_0 e^{\beta_0}}{S_0}$ vs. $S_0$] {
    \includegraphics[width=0.47\textwidth]{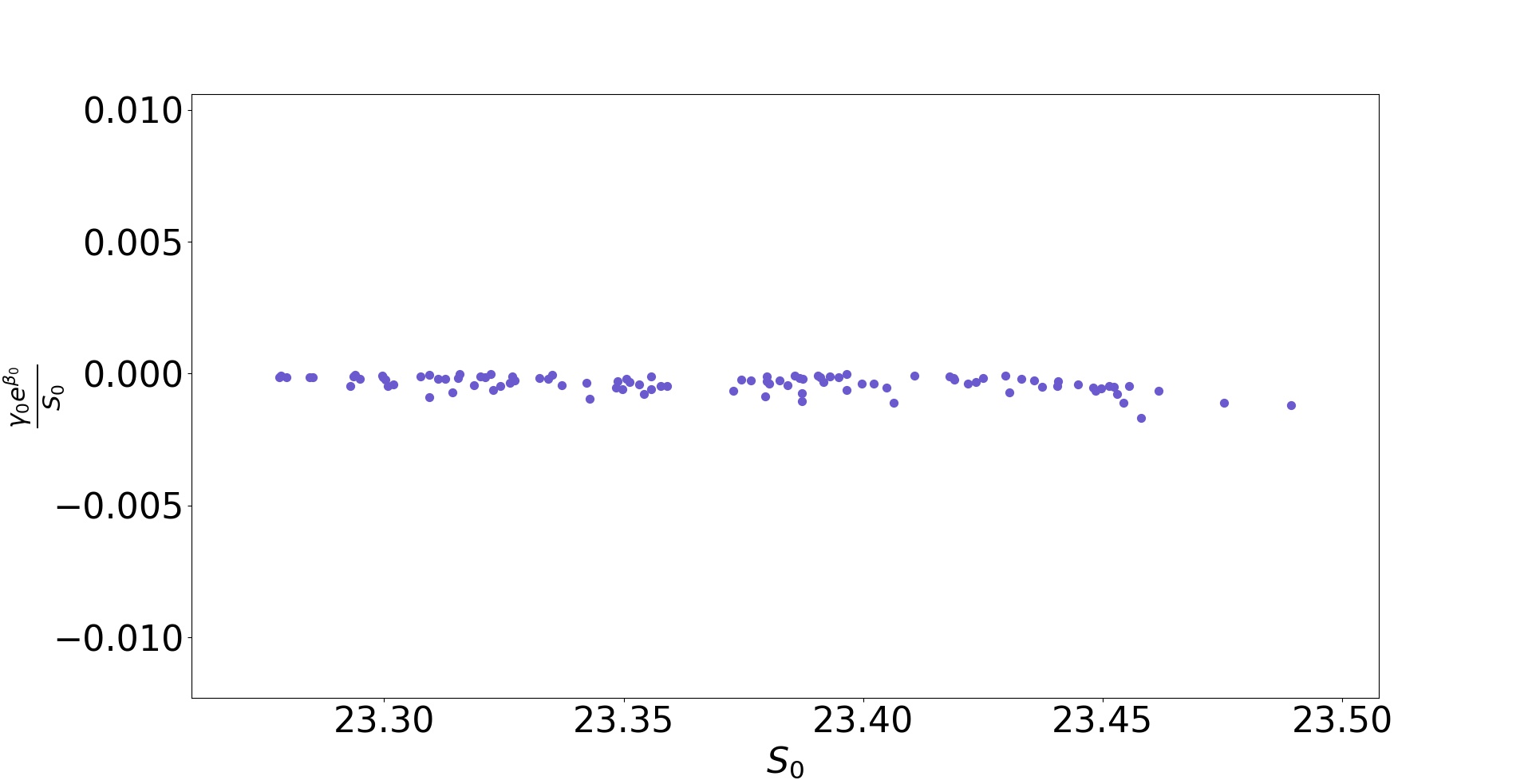}
    \label{tfig1}
  }
  \subfigure[$\frac{\gamma_0 e^{\beta_0}}{S_0}$ vs. $\alpha_0$] {
    \includegraphics[width=0.47\textwidth]{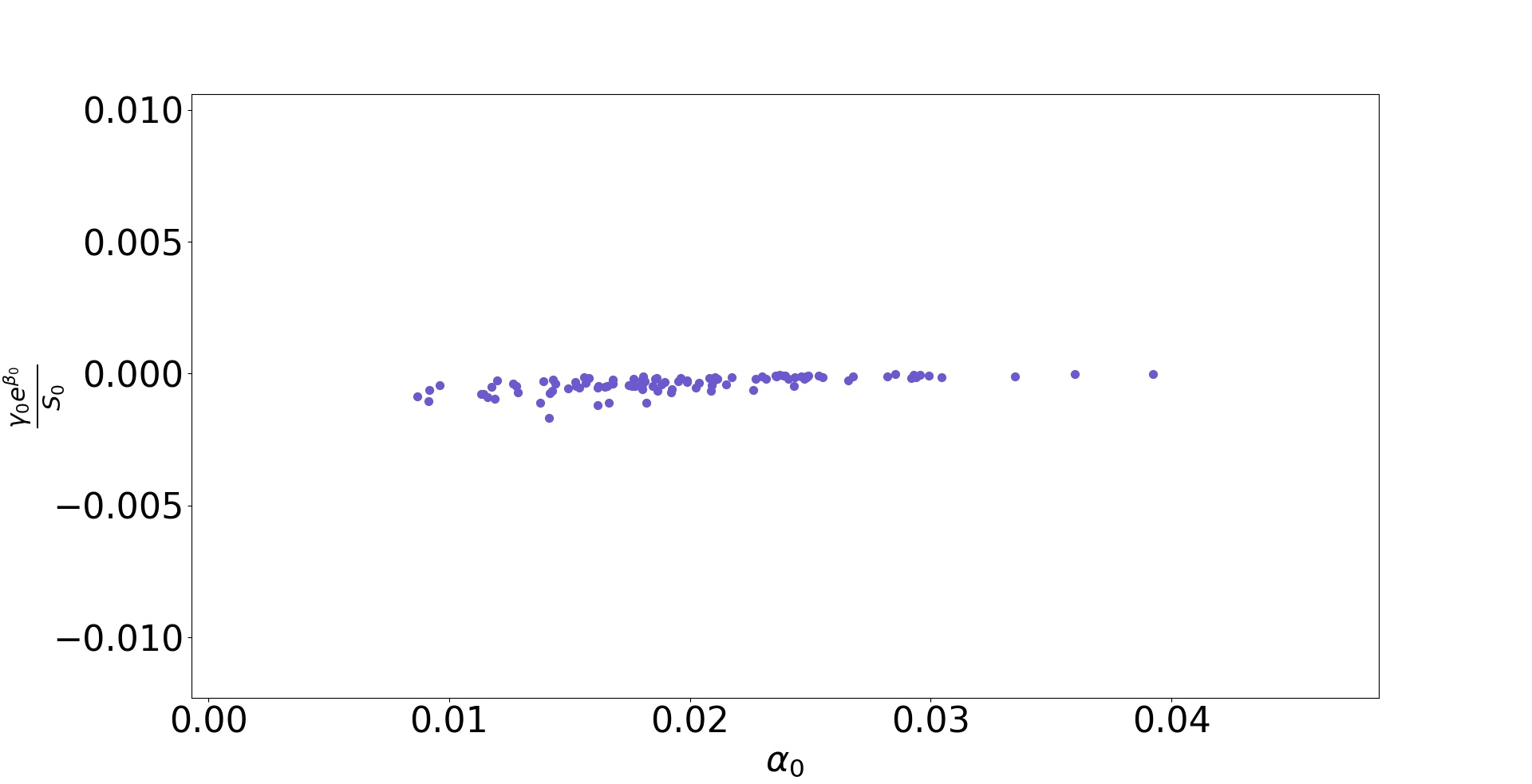}
    \label{tfig2}
  }
  \subfigure[$\alpha_0$ vs. $S_0$] {
    \includegraphics[width=0.47\textwidth]{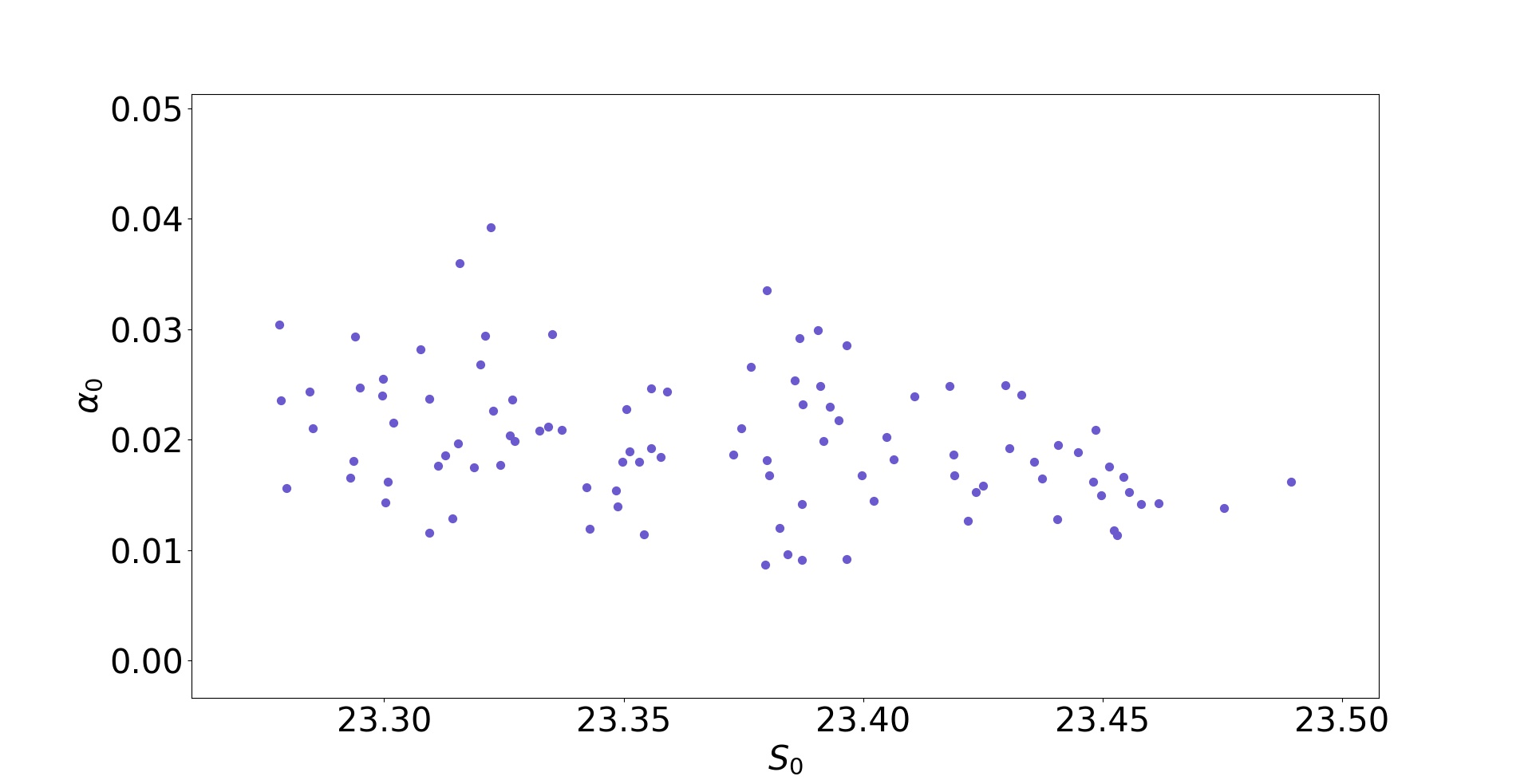}
    \label{tfig3}
  }
  \caption{Correlation scatter plots of of the extrinsic parameters from a simplified form of Equation~\ref{equation_turbofan}, i.e. $S = S_0(1 + \frac{\gamma_0 e^{\beta_0}}{S_0} e^{\alpha_0 c})$. }
  \label{turbofan_parameters}
\end{figure}

Figure~\ref{turbofan_metric} shows results on some structural form candidates found with low values for the $L$ metric.
Using the approach as described in this work, we found the general equation that best describes the dynamics of the engines as 
\begin{equation}
\label{equation_turbofan}
S = \gamma_0 e^{\alpha_0 c + \beta_0} + S_0,
\end{equation}
where $S$ is the sensor channel measurement, and $c$ is the working cycle of the engine.
Figure~\ref{engines} shows the data for different turbofan engines (test set) and their corresponding best fitted curves. 
In Figure~\ref{turbofan_parameters}, we show correlation scatter plots of the irreducible extrinsic parameters from Equation~\ref{equation_turbofan}. 
From Figure~\ref{turbofan_parameters}, one notice that there are two main parameters that uniquely describe each system, i.e. $\alpha_0$ and $S_0$.

\subsection {Case: influence of a parameter on the behavior of a system}
\label{caseconcrete}

In the previous illustration in Section~\ref{nasa}, we neither have the information nor control on the system properties of the various engines which made each engine to be different from one another. 
If there are some degree of control and information on the system properties, the governing equation can be further determined beyond structural forms. 

Specifically, our work can be further applied to the case when the function is of the form $y = f(x; s)$, where $s$ is a system property that can be varied with some degree of control. In such a case, the extrinsic parameters $\theta_i$ can only be at most, a function of $s$. By repeating the approach as described in this work on $\theta = \theta(s)$, one can fully determine $f(x; s)$.

We illustrate the procedure with an example application: the trajectory of a system (projectiles, for example plant seeds) that behaves according to
\begin{equation}
\label{projectile}
y = \Bigl(\frac{\sec \theta}{G} + \tan \theta\Bigl) x + \ln\Bigl(1 - \frac{x \sec \theta}{G}\Bigl),
\end{equation}
where $x$ and $y$ are the x- and y- positions of the system, and $G$ is a controlled system property that characterizes the shape and mass of the system while $\theta$ is fixed to be a constant for all systems.
Details on Equation~\ref{projectile} for trajectories of systems with property $G$ can be found in \cite{Ribeiro2018ProjectileMT,Stewart_2011}.

Of course, the experimentalist does not know \textit{a priori} Equation~\ref{projectile}, and is tasked to discover the equation based on the collected datasets from multiple systems with different $G$ values. 
For the purpose of this illustration, we simulate datasets with different $G$ values, and then applied our approach to the datasets. 

\begin{figure}
  \centering
  \subfigure[] {
    \includegraphics[width=1.\textwidth]{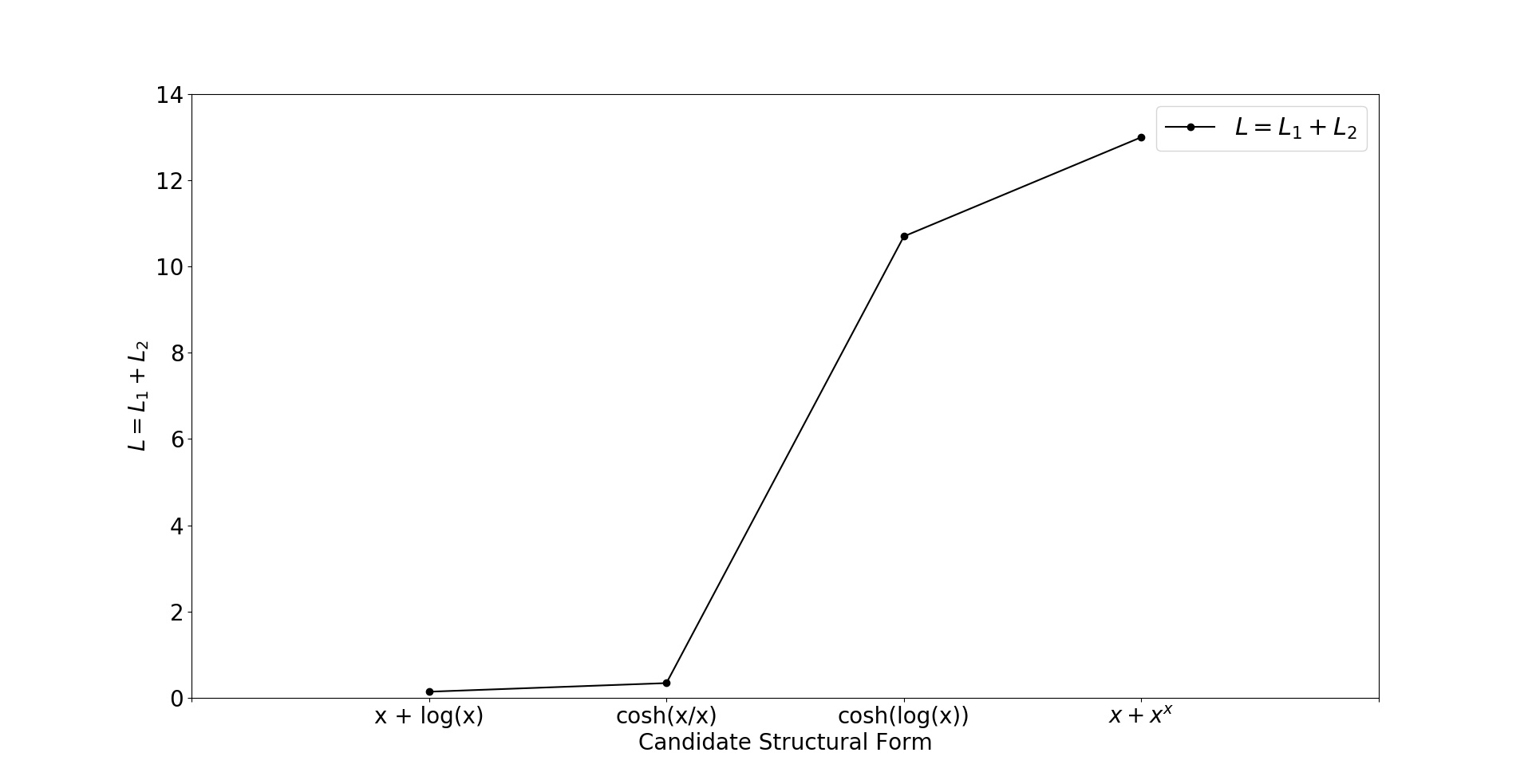}
    \label{fig_1s}
  }
  \subfigure[] {
    \includegraphics[width=1.\textwidth]{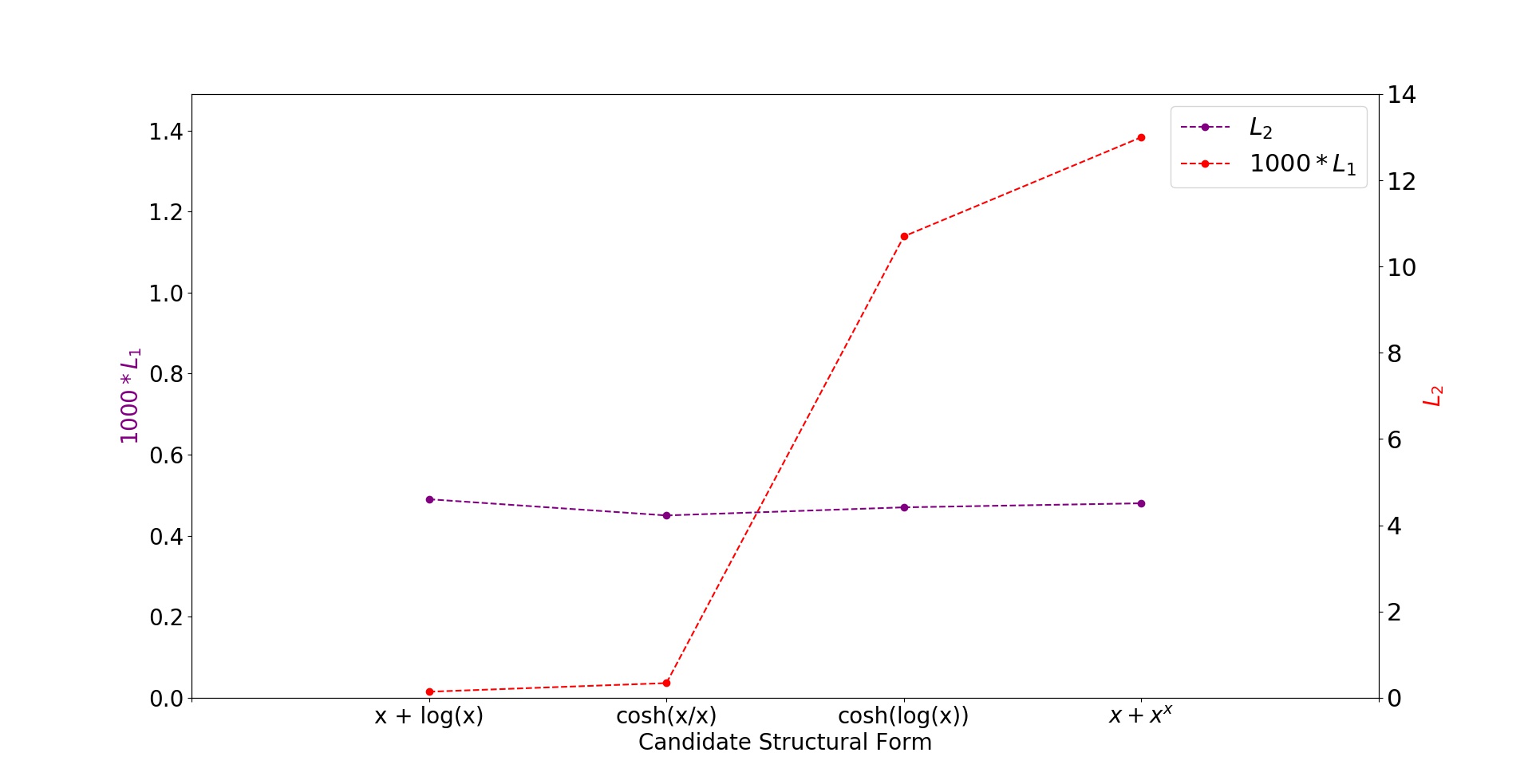}
    \label{fig_2s}
  }
  \caption{Structural form candidates (before transformation) with low values for the $L$ metric (see Figure (a)). $L_1$ and $L_2$ (see Figure (b)) are the values for the first and second terms in Equation~\ref{simplemstar}.}
  \label{seed_metric}
\end{figure}

\begin{figure}
  \centering
  \subfigure[] {
    \includegraphics[width=0.47\textwidth]{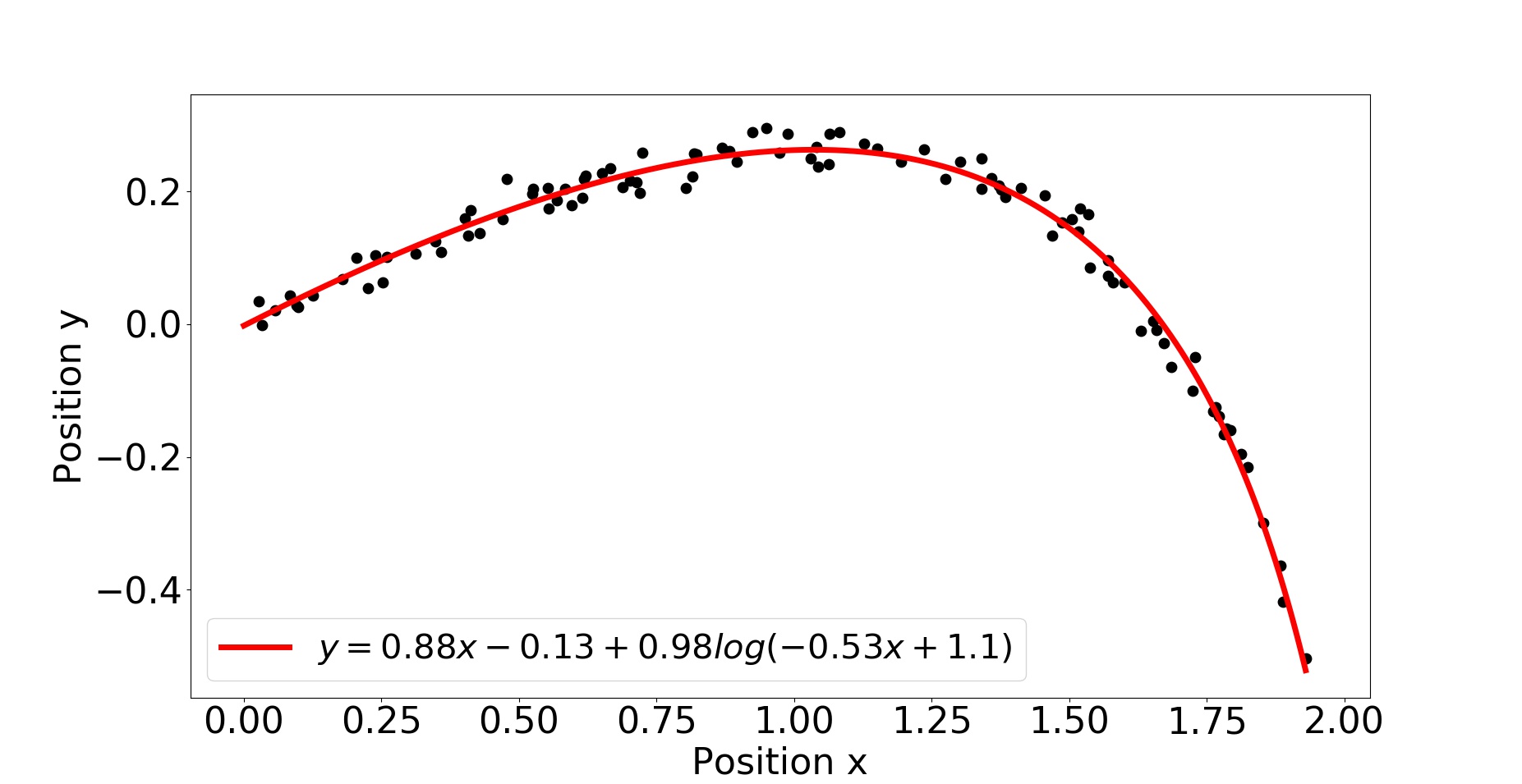}
    \label{fig1}
  }
  \subfigure[] {
    \includegraphics[width=0.47\textwidth]{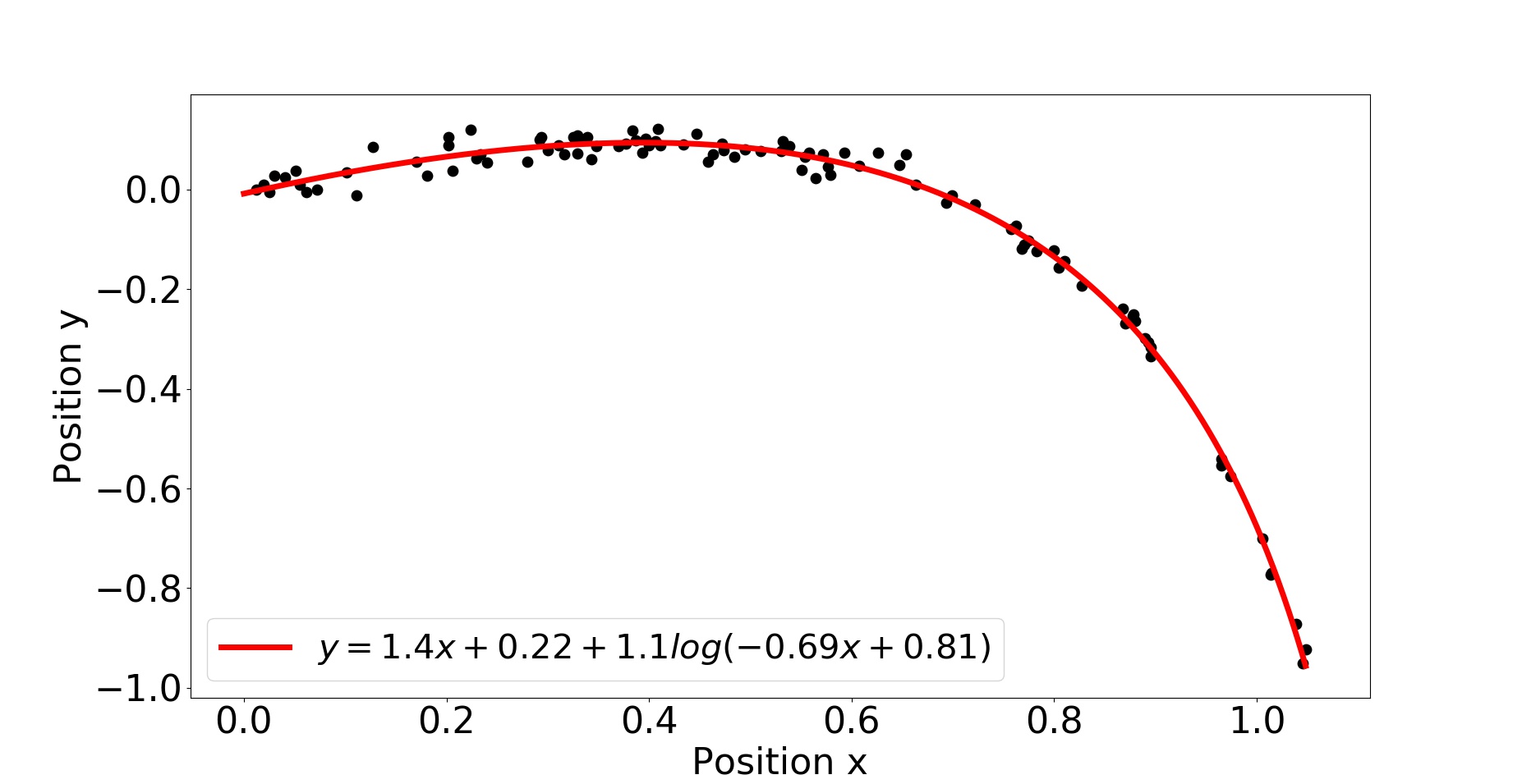}
    \label{fig2}
  }
  \subfigure[] {
    \includegraphics[width=0.47\textwidth]{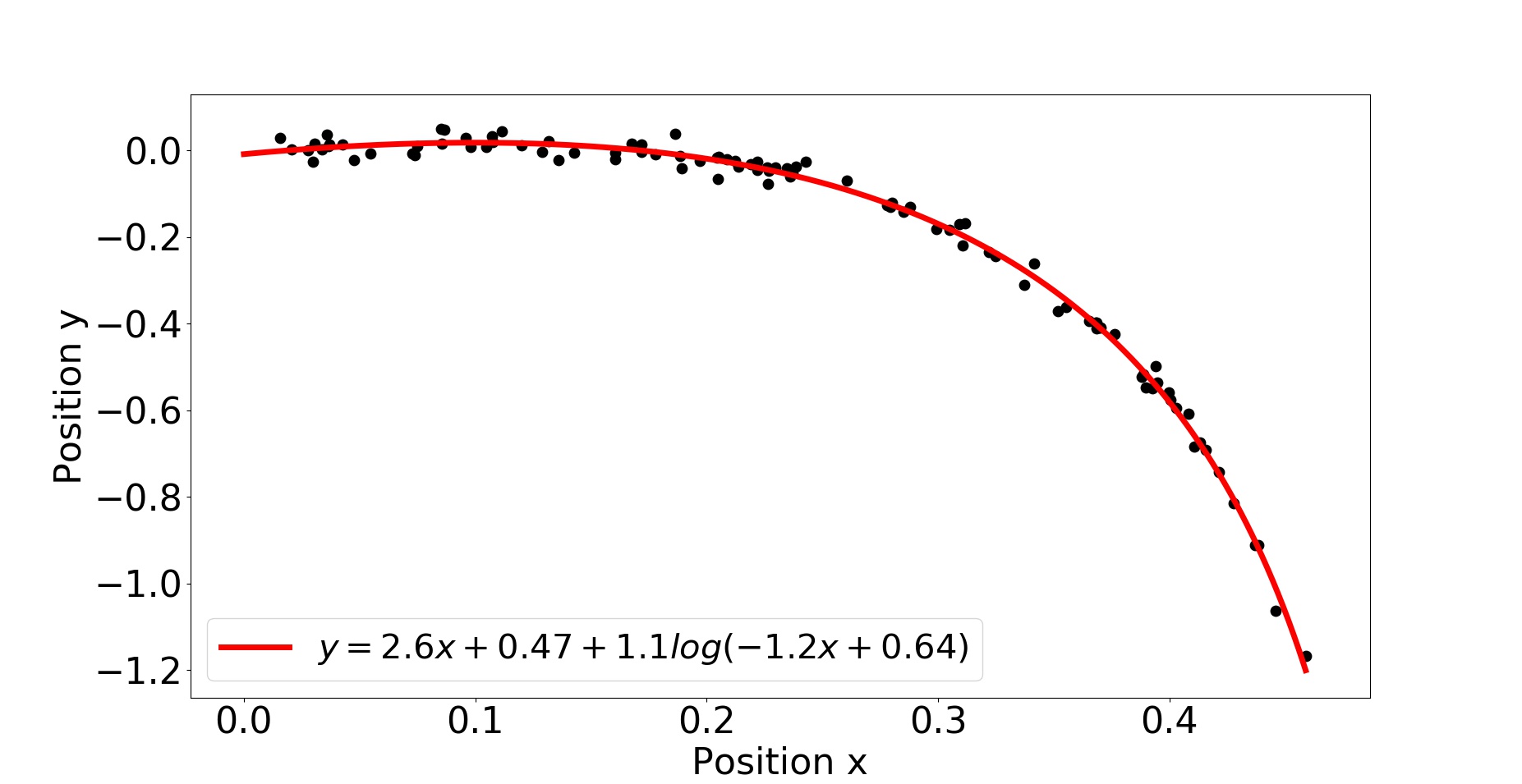}
    \label{fig3}
  }
  \subfigure[] {
    \includegraphics[width=0.47\textwidth]{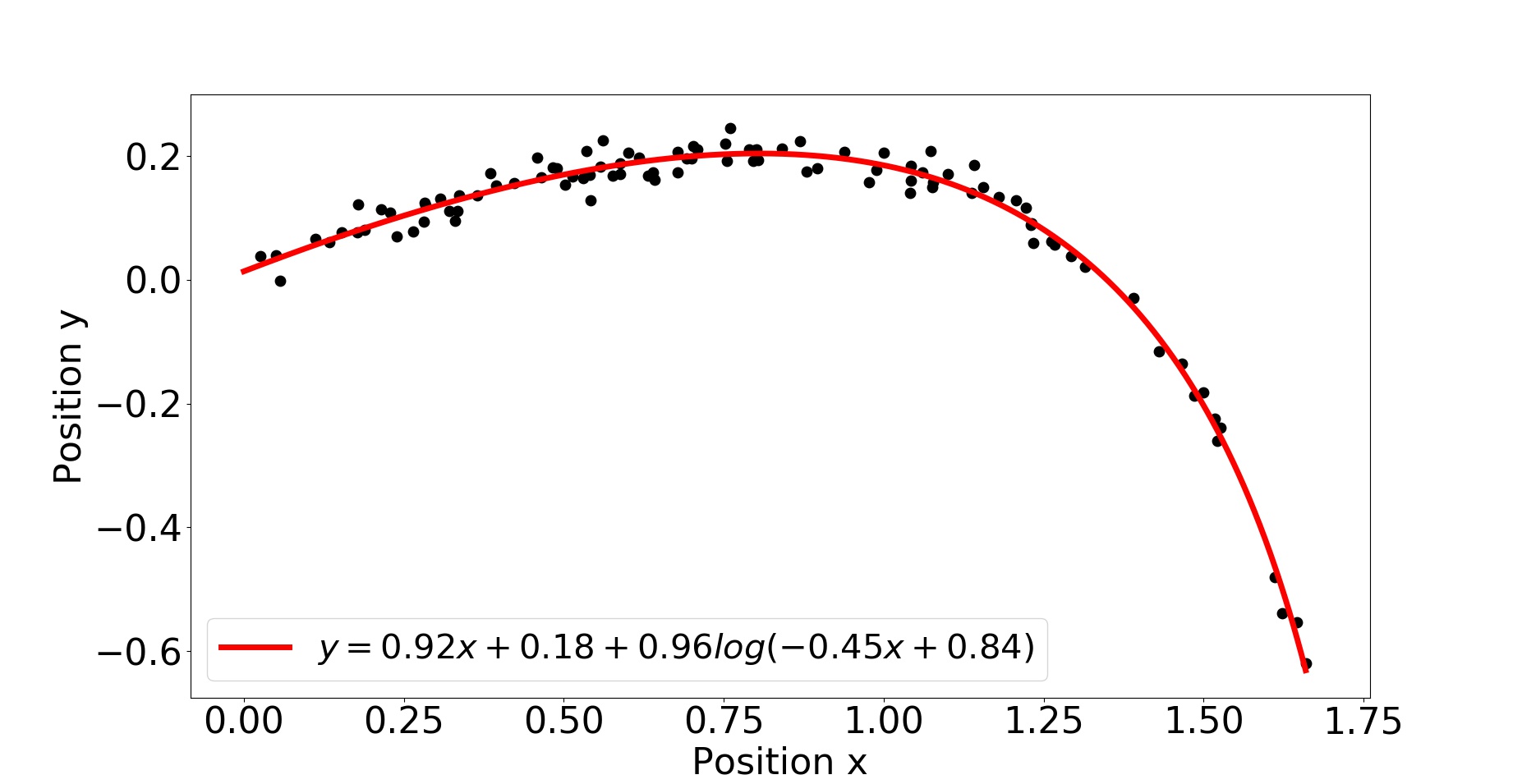}
    \label{fig4}
  }
  \caption{Data points of four examples and their fitted curves using the found structural form of the general equation. In each fitted equation, the constants $\beta_0$ and $y_0$ are reported as a combined value. }
  \label{concrete}
\end{figure}

Figure~\ref{seed_metric} shows results on some structural form candidates found with low values for the $L$ metric.
The equation found to be the best that describes the datasets is as expected: 
\begin{equation}
\label{equation_concrete}
y = (\alpha_0 x + \beta_0) + \gamma_0 \ln(\alpha_1 x + \beta_1) + y_0.
\end{equation}
Figure~\ref{concrete} shows the data of four example systems associated with different $G$ values and the corresponding fitted curves of $y$ vs. $x$.

%
%

\begin{figure}
  \centering
  \subfigure[$\alpha_0$ vs. $G$] {
    \includegraphics[width=0.47\textwidth]{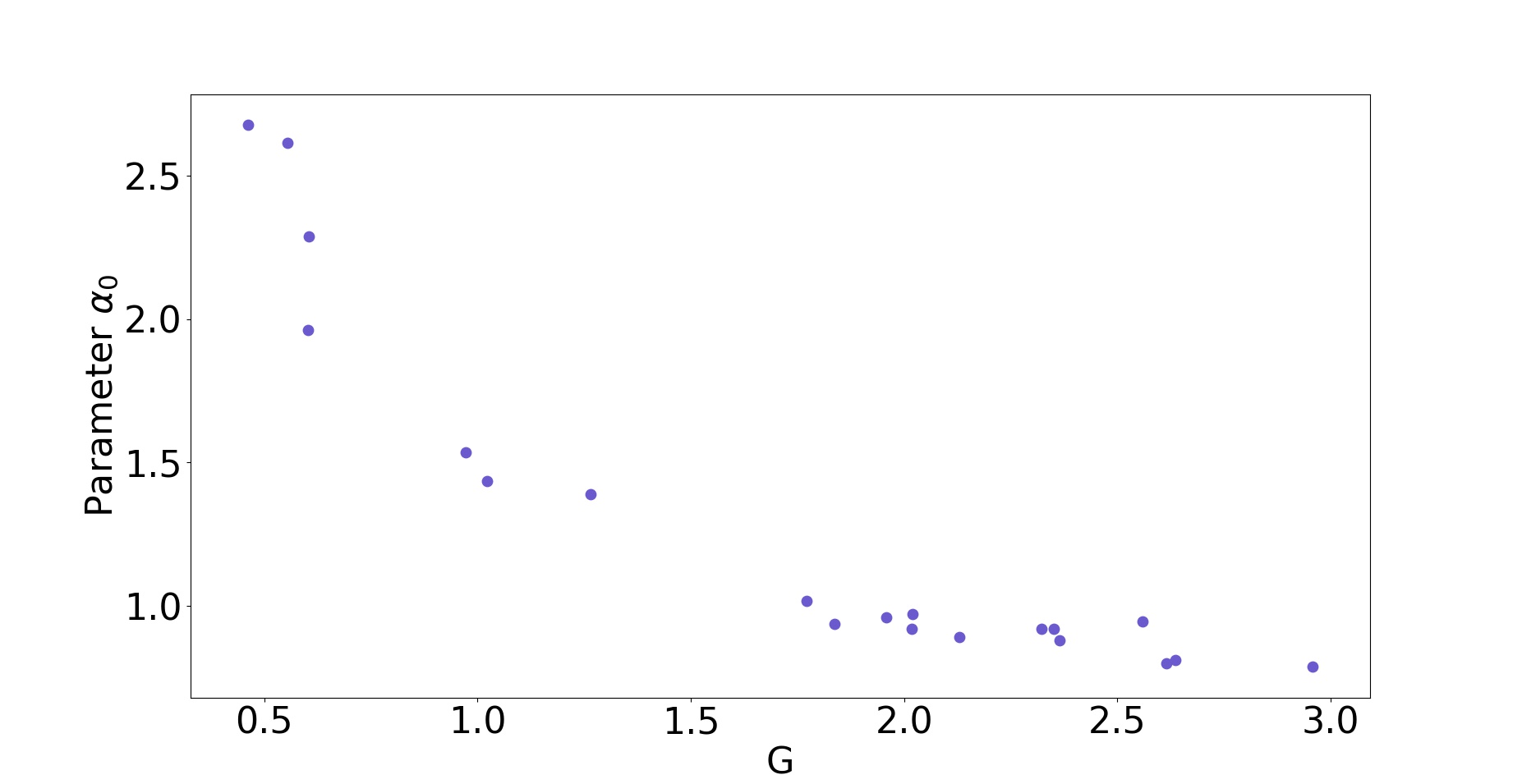}
    \label{fig1}
  }
  \subfigure[$\gamma_0$ vs. G] {
    \includegraphics[width=0.47\textwidth]{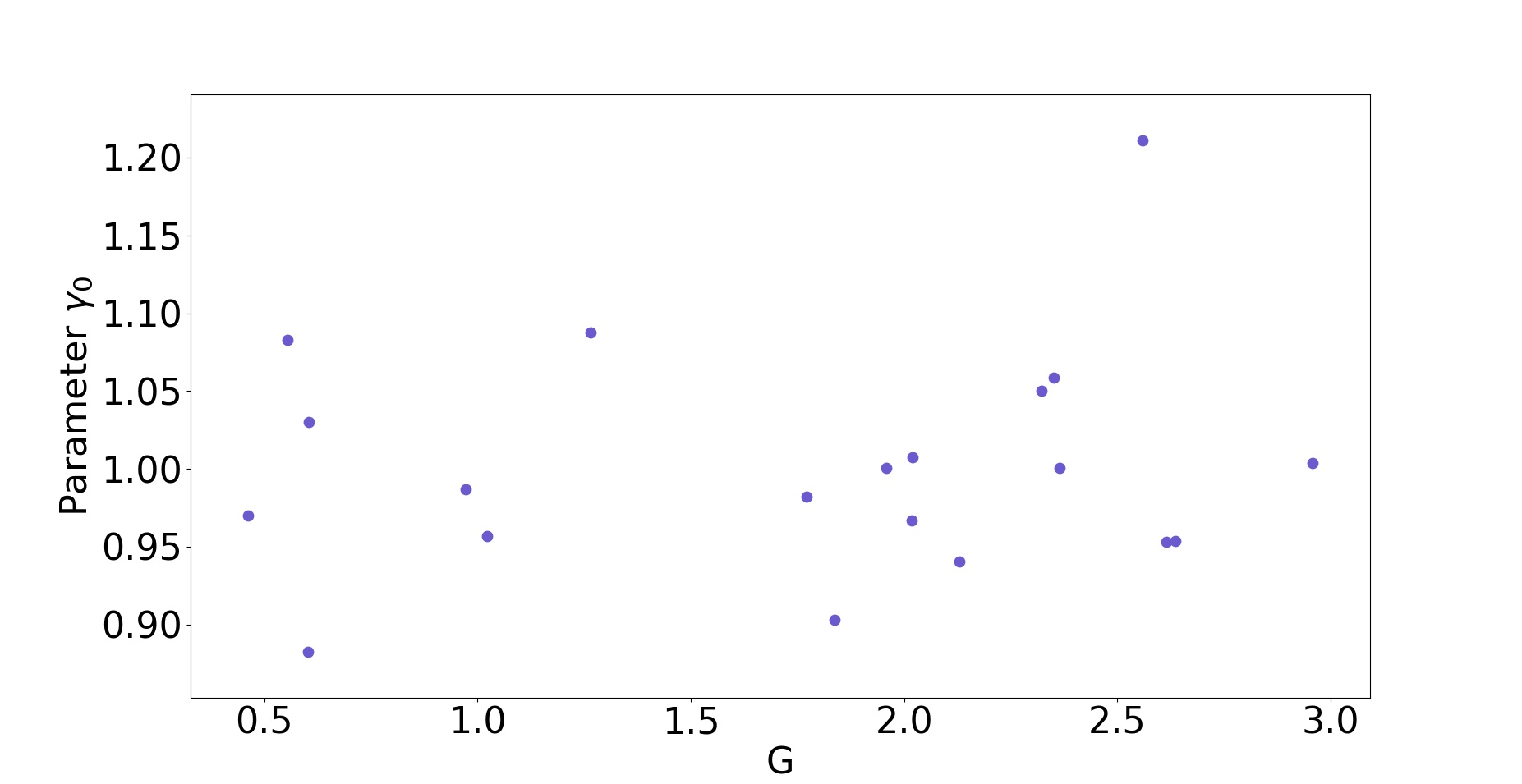}
    \label{fig2}
  }
  \subfigure[$\alpha_1 e^\frac{\beta_0 + y_0}{\gamma_0}$ vs. G] {
    \includegraphics[width=0.47\textwidth]{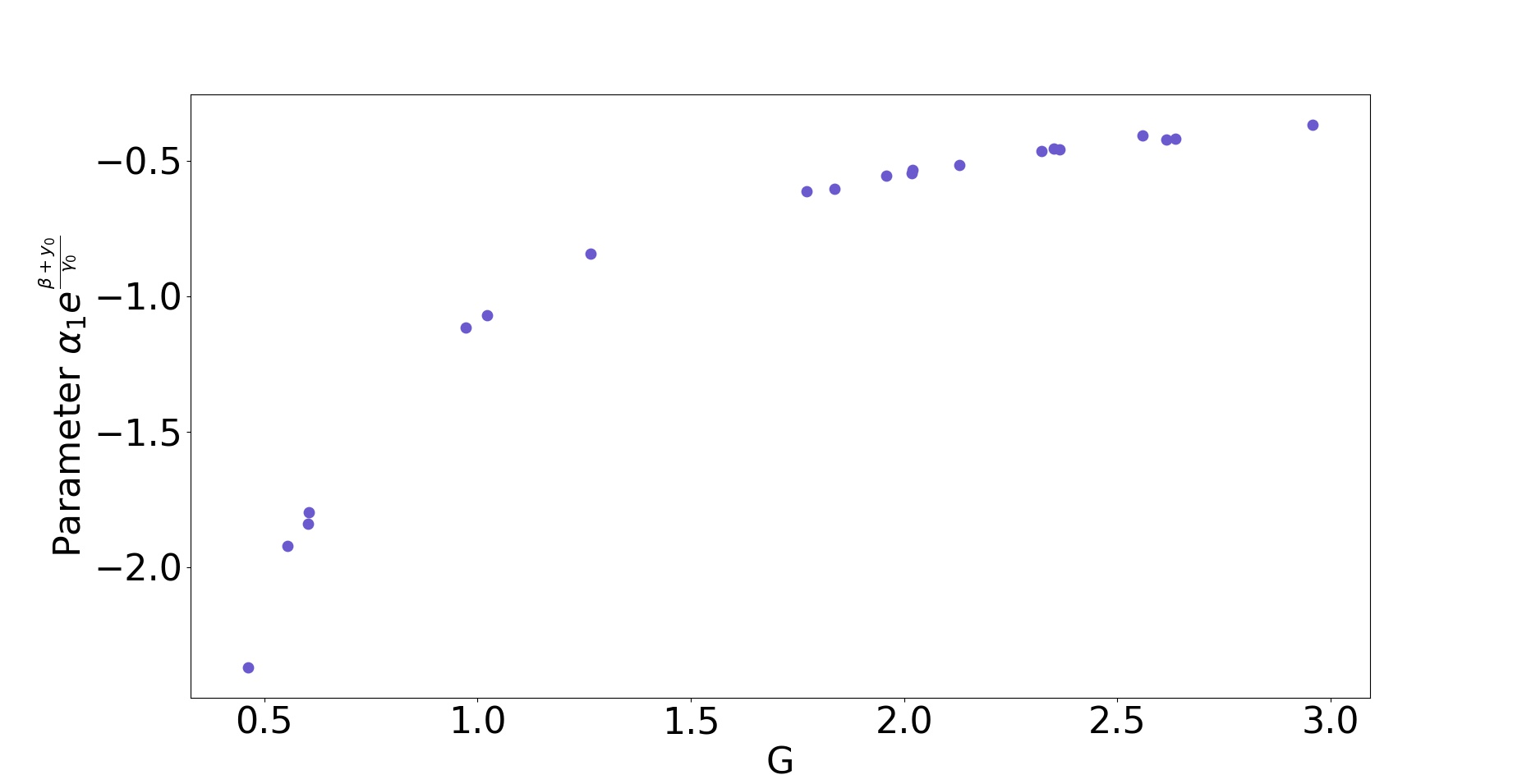}
    \label{fig3}
  }
  \subfigure[$\beta_1 e^\frac{\beta_0 + y_0}{\gamma_0}$ vs. G] {
    \includegraphics[width=0.47\textwidth]{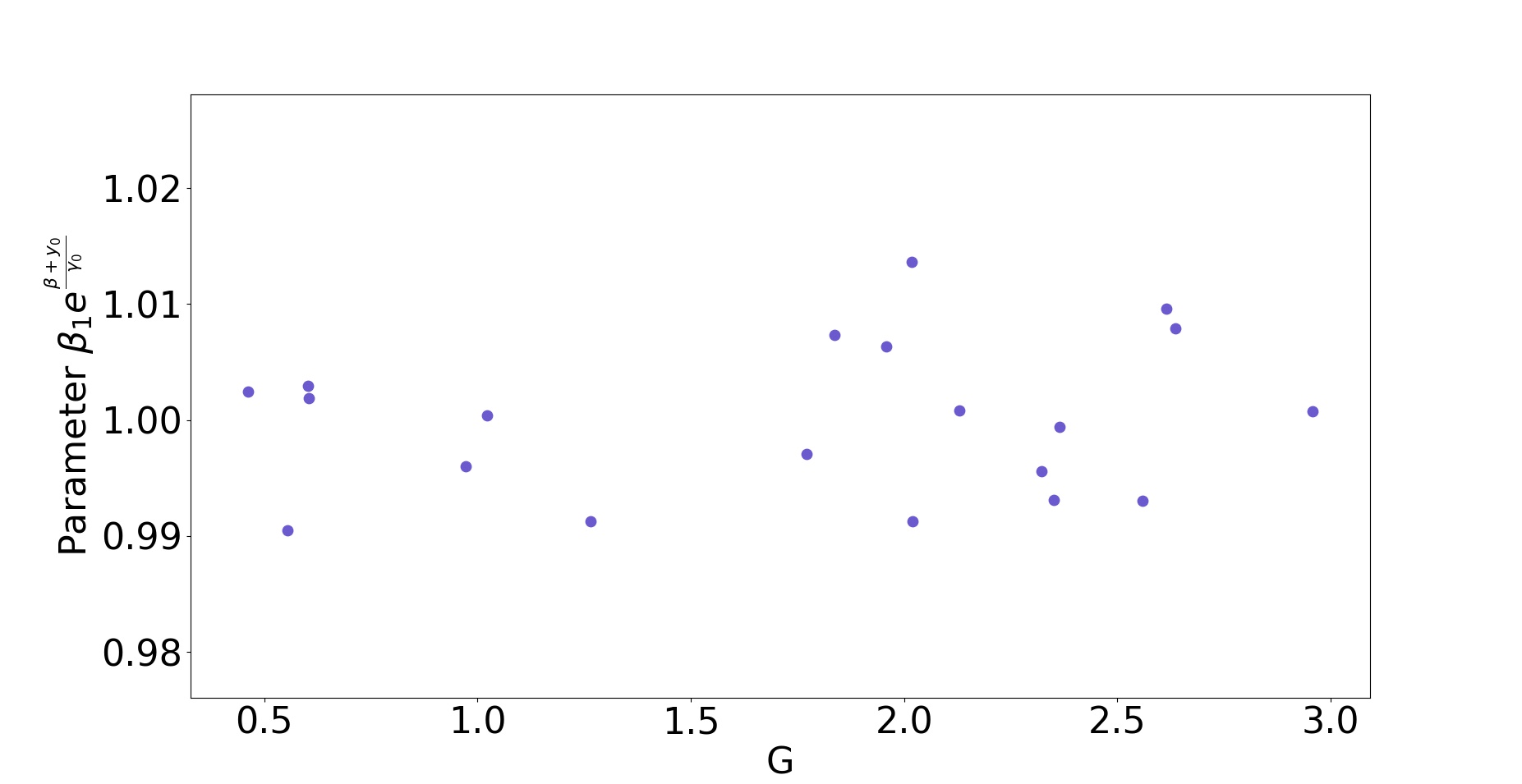}
    \label{fig4}
  }
  \caption{Scatter plots of the values of the fitted parameters.}
  \label{parameters}
\end{figure}

In Figure~\ref{parameters}, we show the curves of irreducible extrinsic parameters from Equation~\ref{equation_concrete}. One can repeat the same approach and perform further fits on such curves, hence finding the relationship between the extrinsic parameters and the known to be varying intrinsic parameter $G$. In this manner, this illustration shows how one can completely determine the model, not only its structural form that explains the data.

\section{Summary and Future Work}

In summary, we have shown a general way of employing symbolic regression on measurement data from multiple but similar systems; in order to identify the underlying structural form of the general mathematical relationship governing their dynamics, by leveraging on their collective information. To this end, we proposed a set of transformation rules and an $L$ metric revolving around some arguments based on the nature of physical systems. The benefits of being able to find the structural form is that it alleviates the need for retraining of machine learning models for a new similar system. One can simply do a curve fit of the structural form to some data collected from that system of interest to predict its future behavior. Indeed, the structural form can be regarded as a white-box model instead of a black-box usually associated with machine learning models.

In general, there can be many models that adequately explains a dataset, as is also the case in our two illustrative examples, where there are various models with low $MSE$ w.r.t. the data. However, the strength of the interplay between the extrinsic parameters in each model can vary tremendously. It is this strength that we argue helps identify the best model w.r.t. the data.
In one of the illustrative case, the hidden system properties have limited how far one can completely determine a model to explain the data. Our future work will revolve around understanding the interplay between these hidden properties, and how they would reveal itself in the form of extrinsic parameters.




\vspace*{-0.01in}
\noindent
\rule{12.6cm}{.1mm}


\begin{thebibliography}{0}



\bibitem{CHEN20181973}C. Chen, C. Luo and Z. Jiang,
"Block building programming for symbolic regression,"
{\it Neurcomputing} {\bf 275} (2018) 1973--1980.

\bibitem{inproceedings}D. V. Chigirev and W. Bialek,
"Optimal manifold representation of data: an information theoretic approach,"
{\it Adv. in Neural Info. Processing Sys. 16: Proceedings of the 2003 Conf.} (2003).

\bibitem{Koza}J. Koza, 
{\it Genetic Programming: On the Programming of Computers by Means of Natural Selection},
MIT Press, Cambridge, 1992.

\bibitem{BV}S. Geman, E. Bienenstock and R. Doursat,
"Neural networks and the bias/variance dilemma,"
{\it Neural Computation} {\bf 4} (1992) 1--58.

\bibitem{Grnwald2004ATI}P. Gr{\"u}nwald,
"A tutorial introduction to the minimum description length principle,"
{\it Arxiv:math.ST/0406077} (2004).

\bibitem{MDL}P. Gr{\"u}nwald, I. J. Myung and M. A. Pitt,
{\it Advances in minimum descriptioni length: theory and applications},
MIT Press, Cambridge, 2005.

\bibitem{FRENAY20131}B. Frenay, G. Doquire and M. Verleysen,
"Is mutual information adequate for feature selection in regression?"
{\it Neural Net.} {\bf 48} (2013) 1--7.

\bibitem{Goodfellow-et-al-2016}I. Goodfellow, Y. Bengio and A. Courville,
{\it Deep Learning}, MIT Press, Cambridge, 2016.

\bibitem{Lei2018GeometricUO}N. Lei, Z. Luo, S. T. Yau and X. Gu,
"Geometric understanding of deep learning,"
{\it Arxiv:1805.10451} (2018).

\bibitem{Lin2015}B. Lin, 
"A geometric viewpoint of manifold learning,"
{\it Applied Informatics} {\bf 2} (2015) 3.

\bibitem{doi:10.1002/widm.8}W. Y. Loh,
"Classification and regression trees,"
{\it Wiley Interdisciplinary Revs: Data Mining and Knowledge Discovery} {\bf 1} (2011) 14--23.

\bibitem{MARINI2015153}F. Marini and B. Walczak,
"Particle swarm optimization (pso). a tutorial,"
{\it Chem. and Intelli. Lab. Sys. } {\bf 149} (2015) 153--165.

\bibitem{ecology}B. Martin, S. Much and A. M. Hein,
"Reverse-engineering ecological theory from data,"
{\it Proc. of the Royal Soc. B: Bio. Sci.} {\bf 285} (2018) 20180422.

\bibitem{Mwanje_1980}J. Mwanje,
"An approach to Hooke's law,"
{\it Phys. Edu.} {\bf 15} (1980) 2.

\bibitem{Ribeiro2018ProjectileMT}W. J. M. Ribeiro and J. R. de Sousa,
"Projectile motion: the "coming and going" phenomenon,"
{\it Arxiv:1805.08066} (2018).

\bibitem{NASA}A. Saxena and K. Goebel,
"Turbofan engine degradation simulation data set," (2008).

\bibitem{Schmidt81}M. Schmidt and H. Lipson,
"Distilling free-form natural laws from experimental data,"
{\it Science} {\bf 324} (2009) 81--85.

\bibitem{inbook} G. Smits and M. Kotanchek,
``Pareto-front exploitation in symbolic regression,''
{\it Genetic Programming Theory and Practice II}, eds.~U. O'Reilly,
T. Yu, R. Riolo and B. Worzel, Springer, 2006,
pp.~283--299.


\bibitem{Stewart_2011}S. M. Stewart, 
"On the trajectories of projectiles depicted in the early ballistic woodcuts,"
{\it Eur. J. of Phys.} {\bf 33} (2011) 149-166.

\bibitem{toush}R. Toushmalani, Z. Parsa and A. Esmaeili,
"Comparison result of inversion of gravity data of a fault by cuckoo optimization and levenberg-marquardt methods,"
{\it Res. J. of Pharm., Bio. and Chem. Sci.} {\bf 5} (2014) 418--427.


\bibitem{4632147}E. J. Vladislavleva, G. F. Smits and D. den Hertog,
"Order of nonlinearity as a complexity measure for models generated by symbolic regression via pareto genetic programming,"
{\it IEEE Trans. on Evol. Comput.} {\bf 13} (2009) 333--349.

\bibitem{Weiss2016}K. Weiss, T. M. Khoshgoftaar and D. Wang,
"A survey of transfer learning,"
{\it J. of Big Data} {\bf 1} (2016).

\bibitem{10.1007/978-3-642-27549-4_34}A. C. Z{\u{a}}voianu, G. Kronberger, M. Kommend, D. Zaharie and M. Affenzeller,
"Improving the parsimonisity of regression models for and enhanced genetic programming process,"
{\it Computer Aided Systems Theory -- EUROCAST 2011}, eds.~R. Moreno-D{\'i}az, F. Pichler, A. Quesada-Arencibia,
Springer, 2012, pp.~264--271. 

\end{thebibliography}
\end{document}